\documentclass[10pt,letterpaper,oneside,onecolumn,journal,draftcls]{IEEEtran}
\usepackage{subfigure,epsfig}
\newcommand{\figwidth}{1.3in}
\newcommand{\smallfigwidth}{1.0in}
\newcommand{\zsmallfigwidth}{1.1in}
\newcommand{\graphwidth}{4.0in}
\begin{document}
\title{Towards the Evolution of Novel Vertical-Axis \\ Wind Turbines}
\author{Richard~J.~Preen and Larry~Bull%
\thanks{The authors are with the Department of Computer Science and Creative Technologies, University of the West of England, Bristol BS16 1QY, U.K. (e-mail: richard.preen@uwe.ac.uk, larry.bull@uwe.ac.uk).}%
\thanks{}}
\markboth{}
{PREEN \MakeLowercase{and} BULL: TOWARDS THE EVOLUTION OF NOVEL VERTICAL-AXIS WIND TURBINES}
\maketitle
\begin{abstract}
Renewable and sustainable energy is one of the most important challenges currently facing mankind. Wind has made an increasing contribution to the world's energy supply mix, but still remains a long way from reaching its full potential. In this paper, we investigate the use of artificial evolution to design vertical-axis wind turbine prototypes that are physically instantiated and evaluated under approximated wind tunnel conditions. An artificial neural network is used as a surrogate model to assist learning and found to reduce the number of fabrications required to reach a higher aerodynamic efficiency, resulting in an important cost reduction. Unlike in other approaches, such as computational fluid dynamics simulations, no mathematical formulations are used and no model assumptions are made.
\end{abstract}
\section{Introduction}
\IEEEPARstart{I}{n recent} years, wind has made an increasing contribution to the world's energy supply mix. However, there is still much to be done in all areas of the technology for it to reach its full potential. Currently, horizontal-axis wind turbines (HAWTs) are the most commonly used form. However, ``modern wind farms comprised of HAWTs require significant land resources to separate each wind turbine from the adjacent turbine wakes. This aerodynamic constraint limits the amount of power that can be extracted from a given wind farm footprint. The resulting inefficiency of HAWT farms is currently compensated by using taller wind turbines to access greater wind resources at high altitudes, but this solution comes at the expense of higher engineering costs and greater visual, acoustic, radar and environmental impact'' \cite{Dabiri:2011}. This has forced wind energy systems away from high energy demand population centres and towards remote locations with higher distribution costs. In contrast, vertical-axis wind turbines (VAWTs) do not need to be oriented to wind direction and can be positioned closely together, potentially resulting in much higher efficiency. VAWT can also be easier to manufacture, may scale more easily, are typically inherently light-weight with little or no noise pollution, and are more able to tolerate extreme weather conditions (see, e.g., \cite{Eriksson:2008} for discussions). However, their design space is complex and relatively unexplored. Generally, two classes of design are currently under investigation and exploitation: the Savonius, which has blades attached directly upon the central axis structure; and the Darrieus, where the blades---either straight or curved---are positioned predominantly away from the central structure. Hybrids also exist.

The majority of blade design optimisation is performed through the use of computational fluid dynamics (CFD) simulations, typically described with three-dimensional Navier-Stokes equations \cite{Anderson:1995}. However, three-dimensional CFD simulations are computationally expensive, with a single calculation taking hours on a high-performance computer, making their use with an iterative search approach difficult \cite{Graning:2007}. Moreover, assumptions need to be made, e.g., regarding turbulence or pressure distributions, which can significantly affect accuracy. Previous evolutionary studies have been undertaken with types of CFD to optimise the blade profile for both HAWT (e.g., \cite{Hampsey:2002}) and VAWT (e.g., \cite{Carrigan:2012}) to varying degrees of success/realism.
 
Evolutionary algorithms (EAs), as recently highlighted \cite{Koza:2010}, have been used to produce over seventy examples of human-competitive performance. That is, cases where evolutionary computation has produced results which, for example, match or improve upon previously patented designs, match or improve upon the current scientific knowledge, or have solved a problem of indisputable difficulty in its field. EAs have also been used to design three-dimensional physical objects, such as furniture (e.g., \cite{BentleyWakefield:1995}), aircraft engine blades (e.g., \cite{LeeHajela:1996}) and wings (e.g., \cite{OngKeane:2004}). Notably, Lohn {\em et~al.} \cite{Lohn:2008} evolved and manufactured X-band satellite antenna for NASA's ST5 spacecraft, representing the world's first artificially evolved hardware in space. Significantly, the antenna's performance was similar to a design hand-produced by an antenna-contractor. Most of these approaches, however, have used simulations to provide the fitness scores of the evolved designs. 

Embodied evolutionary computing has typically referred to the existence of a physical solution in the fitness evaluation, and can be traced back to the origins of the discipline: the first evolution strategies (ESs) were used to design jet nozzles as a string of real-valued diameters, which were then machined and tested for fitness \cite{Rechenberg:1971}. Other well-known examples include robot controller design (e.g., \cite{Nolfi:1992}), electronic circuit design using programmable hardware (e.g., \cite{Thompson:1998}), product design via human provided fitness values (e.g., \cite{Herdy:1996}), chemical systems (e.g., \cite{Theis:2007}), and unconventional computers (e.g., \cite{HardingMiller:2004}). Evolution in hardware has the potential to benefit from access to a richer environment where it can exploit subtle interactions that can be utilised in unexpected ways. For example, the EA used by Thompson \cite{Thompson:1998} to work with FPGA circuits used some subtle physical properties of the system to solve problems where the properties used are still not understood to this day. Humans can be prevented from designing systems that exploit these subtle and complex physical characteristics through their lack of knowledge, however this does not prevent exploitation through artificial evolution. There is thus a real possibility that evolution in hardware may allow the discovery of new physical effects, which can be harnessed for computation/optimisation \cite{MillerDowning:2002}.  
                                                                                                                                                                                      
Moreover, the advent of high quality, low-cost, additive rapid fabrication technology---known as three-dimensional printing---means it is now possible to fabricate a wide range of prototype designs quickly and cheaply. Three-dimensional printers are now capable of printing an ever growing array of different materials, including sugar (e.g., to help create synthetic livers \cite{Miller:2012}), chemicals (e.g., for custom drug design \cite{Cronin:2012}), cells (e.g., for functional blood vessels \cite{Jakeb:2008}), plastic (e.g., Southampton University laser sintered aircraft), and titanium (e.g., prosthetics such as the synthetic mandible developed by the University of Hasselt and transplanted into an 83-year old woman). Lipson and Pollack \cite{LipsonPollack:2000} were the first to exploit the emerging technology in conjunction with a simulated evolutionary process, printing mobile robots with embodied neural network controllers.

EAs perform a stochastic search for the optimum solution among the design space without the need for any gradient information, however they typically require a large number of fitness evaluations. Techniques to reduce the number of candidate solution evaluations when they are computationally expensive or difficult to obtain/formulate have been developed as evolutionary computation has been applied to more complex domains, e.g., in systems where a human user is involved. This is typically achieved through the construction of models of the problem space via direct sampling---the use of approximations is an established approach in the wider field of optimisation. That is, the evolutionary process uses one or more models to provide the (approximate) utility of candidate solutions, thereby reducing the number of real evaluations during iterations. Initially, all candidate solutions must be evaluated directly on the task to provide rudimentary training data for the modelling, e.g., by neural networks. Periodically, high utility solutions suggested by the model optimisation are then evaluated by the real system. The training data for the model is then augmented with these and the model(s) updated. Over time, as the quality of the model(s) improves, the need to perform real evaluations/fabrications reduces.
 
In this paper, we present results from a pilot study of a learning assisted EA used to design vertical-axis wind turbines wherein prototypes are evaluated under approximated wind tunnel conditions after being physically instantiated by a three-dimensional printer. That is, unlike other approaches, no mathematical formulations are used and no model assumptions are made.
\section{Related Work}
The evolution of geometric models to design arbitrary three-dimensional morphologies has been widely explored. Early examples include Watabe and Okino's lattice deformation approach \cite{WatabeOkino:1993} and McGuire's sequences of polygonal operators \cite{McGuire:1993}. Sims \cite{Sims:1994} evolved the morphology and behaviour of virtual creatures that competed in simulated three-dimensional worlds with a directed graph encoding. Bentley \cite{Bentley:1996} investigated the creation of three-dimensional solid objects via the evolution of both fixed and variable length direct encodings. The objects evolved included tables, heatsinks, penta-prisims, boat hulls, aerodynamic cars, as well as hospital department layouts. Eggenberger \cite{Eggenberger:1997} evolved three-dimensional multicellular organisms with differential gene expression. Jacob and Nazir \cite{JacobNazir:2002} evolved polyhedral objects with a set of functions to manipulate the designs by adding stellating effects, shrinking, truncating, and indenting polygonal shapes. More recently, Jacob and Hushlak \cite{JacobHushlak:2007} used an interactive evolutionary approach with L-systems \cite{PrusinkiewiczLindenmayer:1990} to create virtual sculptures and furniture designs.

EAs have also been applied to aircraft wing design (e.g., \cite{OngKeane:2004}) including aerodynamic transonic aerofoils (e.g., \cite{HaciogluOzkol:2003,QuagliarellaCioppa:1995}), and multidisciplinary blade design (e.g., \cite{HajelaLee:1995}). Few evolved designs, however, have been manufactured into physical objects. Conventionally evolved designs tend to be purely descriptive, specifying what to build but not how it should be built. Thus, there is always an inherent risk of evolving interesting yet unbuildable objects. Moreover, high-fidelity simulations are required to ensure that little difference is observed once the virtual design is physically manifested. In highly complex design domains, such as dynamic objects, the difference between simulation and reality is too large to manufacture designs evolved under a simulator, and in others the simulations are extremely computationally expensive.

Funes and Pollack \cite{FunesPollack:1998} performed one of the earliest attempts to physically instantiate evolved three-dimensional designs by placing physical LEGO bricks according to the schematics of the evolved individuals. A direct encoding of the physical locations of the bricks was used and the fitness was scored using a simulator which predicted the stability of the composed structures. Additionally, Hornby and Pollack \cite{HornbyPollack:2001} used L-systems to evolve furniture designs, which were then manufactured by a three-dimensional printer. They found the generative encoding of L-systems produced designs faster and with higher fitness than a non-generative system. Generative systems are known to produce more compact encodings of solutions and thereby greater scalability than direct approaches (e.g., see \cite{Schoenauer:1996}).

The generative encoding, compositional pattern producing networks \cite{Stanley:2007} have recently been used to evolve three-dimensional objects which were ultimately fabricated on a three-dimensional printer \cite{AuerbachBongard:2010a,AuerbachBongard:2010b,CluneLipson:2011}. Both interactive and target-based approaches were explored.

Recently, Rieffel and Sayles \cite{RieffelSayles:2010} evolved circular two-dimensional shapes where each design was fabricated on a three-dimensional printer before assigning fitness values. Interactive evolution was undertaken wherein the fitness for each printed shape was scored subjectively. Each individual's genotype consisted of twenty linear instructions which directed the printer to perform discrete movements and extrude the material. As a consequence of performing the fitness evaluation in the environment, that is, after manufacture, the system as a whole can exhibit epigenetic traits, where phenotypic characteristics arise from the mechanics of assembly. One such example was found when selecting shapes that most closely resembled the letter `A'. In certain individuals, the cross of the pattern was produced from the print head dragging a thread of material as it moved between different print regions and was not explicitly instructed to do so by the genotype.
\section{Surrogate Assisted Evolution}
The application of EAs can be prohibitive when the evaluations are computationally expensive, an explicit mathematical fitness function is unavailable, or the original fitness function is noisy or multi-modal. Whilst the speed and cost of rapid-prototyping continues to improve, fabricating an evolved design before fitness can be assigned remains an expensive task when potentially thousands of evaluations are required (e.g., 10{\em mins} print time for each very simple individual in \cite{RieffelSayles:2010}). Due to this, a growing body of work is exploring the application of surrogate models (also known as meta-models or response surface models) to provide approximated fitness computations that assist the EA. The use of surrogate models has been shown to reduce the convergence time in evolutionary computation and multiobjective optimisation; see \cite{Jin:2005,Jin:2011,Lim:2010} for recent reviews. Alternative methods, such as fitness inheritance, fitness imitation, and fitness assignment can also be used.

Given a sample $\mathcal{D}$ of evaluated treatments $N$, a surrogate model, $y=f(\vec{x})$, is constructed, where $\vec{x}$ is the genotype, and $y$, fitness, in order to compute the fitness of an unseen data point $\vec{x} \notin \mathcal{D}$. As such, the genotype must be sufficiently compact for the model to optimise. Typically, a set of evaluated genotypes and their real fitness scores are used to perform the supervised training of an MLP-based artificial neural network (e.g., \cite{Bull:1999}); however, other approaches have been explored, e.g., kriging \cite{Ratle:2001}, clustering \cite{KimCho:2001}, support vector regression \cite{Yun:2009}, radial-basis functions \cite{Ong:2006}, and sequential parameter optimisation \cite{Bartz-Beielstein:2006}. The surrogate model is subsequently used to compute estimated fitness values for the EA to utilise. The model must be periodically retrained with new individuals under a controlled evolutionary approach to prevent convergence on local optima. Retraining can be performed by taking either an individual or generational approach \cite{Jin:2000}. In the individual approach, $n$ number of individuals in the population are chosen and evaluated with the real fitness function each generation. In the generational approach, the entire population is evaluated on the real fitness function each $n$-th generation. Resampling methods and surrogate model validation remain an important and ongoing area of research, enabling the comparison and optimisation of models \cite{Bischl:2012}. Both global modelling and local modelling using trust regions (e.g., samples within a certain euclidean distance) are popular approaches.  

Surrogate assisted EAs that use CFD analysis for fitness determination have previously been used to design turbine blades, finding interesting solutions with reduced computational time \cite{Song:2002}. Jin {\em et~al.} \cite{Jin:2002} explored an ES with CFD analysis to minimise the pressure loss of a turbine blade while maintaining a certain outflow angle. The blade representation used consisted of a series of B-spline control points. The population was initialised with a given blade and 2 neural networks were used to approximate the pressure loss and outflow angle, finding faster convergence than without the surrogate models. Gr\"{a}ning {\em et~al.} \cite{Graning:2007} used an ES with covariance matrix adaptation to minimise the pressure loss of a blade using three-dimensional CFD simulations. The ES was augmented by a neural network surrogate model and used a pre-selection resampling approach (where offspring are only generated from individuals evaluated on the real fitness function), however significant improvement over a plain ES was not found.

The surrogate assisted evolution of aerofoil geometries (a type of blade) has been widely explored for use with aircraft design. Some examples include, Giotis and Giannakoglou \cite{GiotisGiannakoglou:1999} who used multiple output neural networks as surrogate models for multiobjective aerofoil optimisation. Emmerich and Naujoks \cite{EmmerichNaujoks:2004} and Kumano {\em et al.} \cite{Kumano:2006} used kriging to provide approximations for multiobjective aerofoil design. In addition, Zhou {\em et~al.} \cite{Zhou:2007} evolved aerodynamic aerofoil geometries with a representation consisting of Hicks-Henne bump function parameters. The EA was assisted by both a global and local surrogate model. Significantly, these approaches use simulations to evaluate candidate solutions and typically consider only two-dimensional blades (due to the cost of CFD analysis).
\section{Methodology}
A vector of 10 integers is here used as a simple and compact encoding of a prototype VAWT. Each allele thus controls 1/10th of a blade. A workspace (maximum object size) of $30\times30\times30mm$ is used so that the instantiated prototype is small enough for timely production ($\sim30mins$) and with low material cost, yet large enough to be sufficient for fitness evaluation. The workspace has a resolution of $100\times100\times100$ voxels. A central platform is constructed for each individual to enable the object to be placed on to the evaluation equipment. The platform consists of a square torus, 1 voxel in width and with a centre of 14 empty voxels. An equilateral cross is then constructed using the genotype, with four blades bent at right angles, resulting in an allele range [1,42]. More specifically, each blade is constructed, starting from the central platform, by enabling the one-tenth of voxels controlled by the allele.

A simple approach to drawing the blades would be to use the allele value to mark the upper position (e.g., allele + centreline) and enable all voxels from that point towards the centreline; where the centreline is a horizontal line at $y$-axis=50 for north-east and south-west quadrants, and a vertical line at $x$-axis=50 for north-west and south-east quadrants. However, to provide more flexibility the following rules are applied. Where the current upper position is greater than or the same as the previous upper position, the voxels are enabled from the current upper position to the previous upper position and extended a further 2 voxels to the centreline (capped at the centreline). If the current upper position is less than or the same as the previous lower position, the voxels are enabled from the current upper position to the previous lower position and extended a further 2 voxels (capped at the maximum grid size). In all other cases, 2 voxels are enabled from the current upper position towards centreline (capped at the centreline). Once the base voxel layer is constructed, it is duplicated to fill the cube in the $z$-dimension. When production is desired, the three-dimensional binary voxel array is converted to stereolithography (STL) format where it may then undergo post-processing before being converted to printer-readable G-code.

\begin{figure}[t]
\centering 
\mbox{\subfigure{\psfig{file=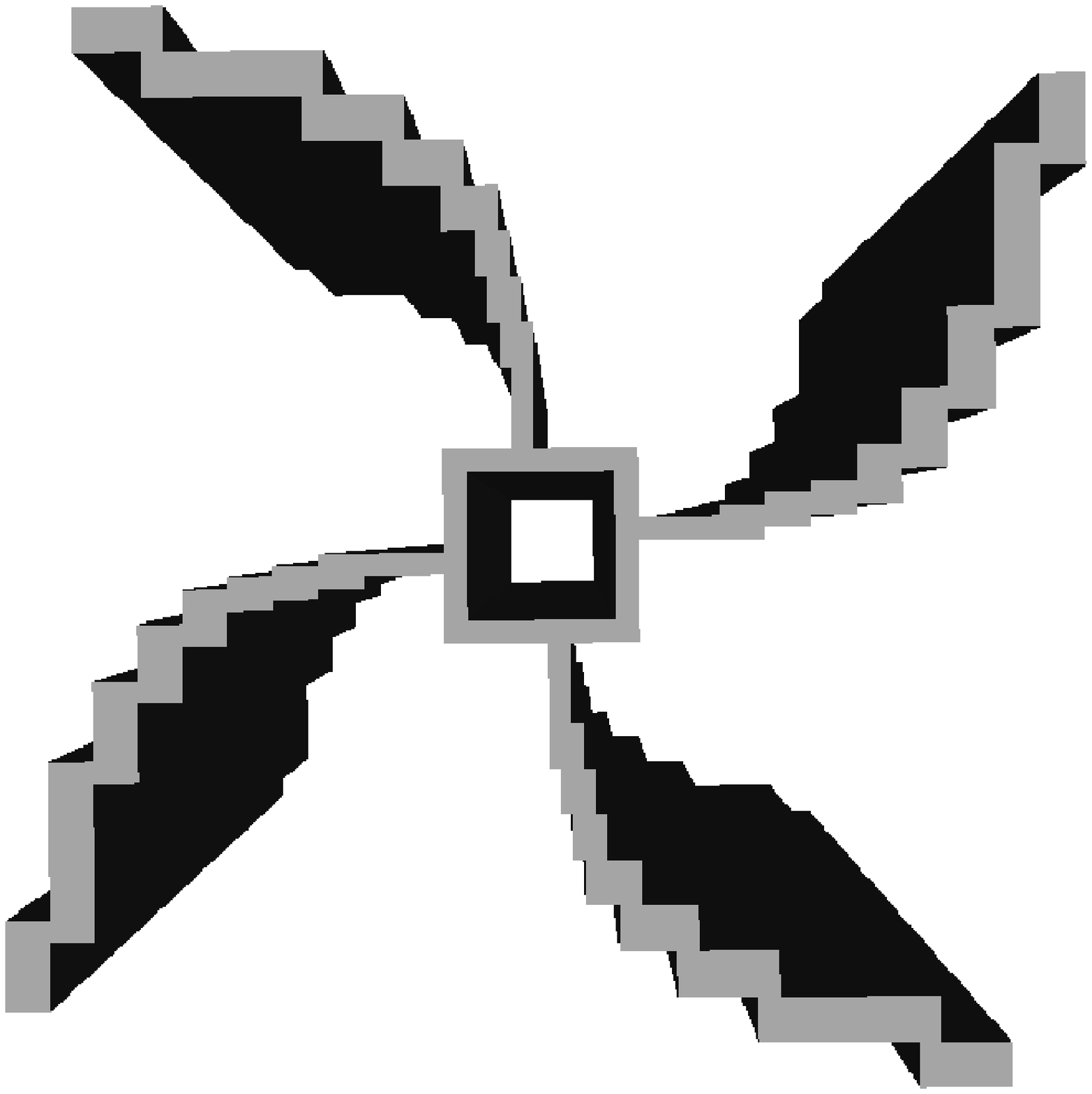,width=\figwidth} } \hspace{0.2in}
\subfigure{\psfig{file=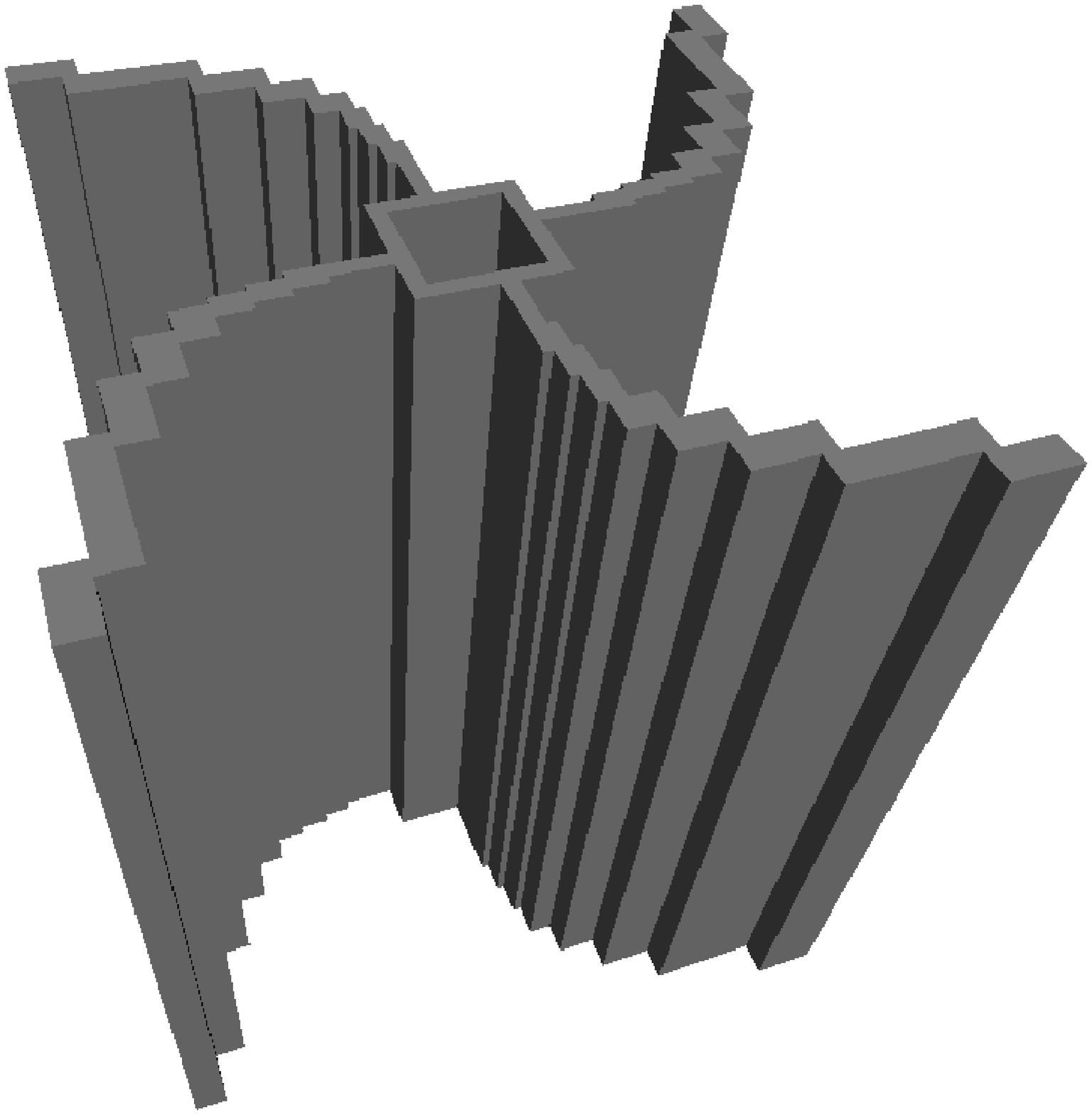,width=\figwidth} }}
\caption{Example phenotype; genome = [2, 2, 3, 4, 5, 8, 13, 20, 34, 40].} \label{fig:target-unsmoothed}
\mbox{\subfigure{\psfig{file=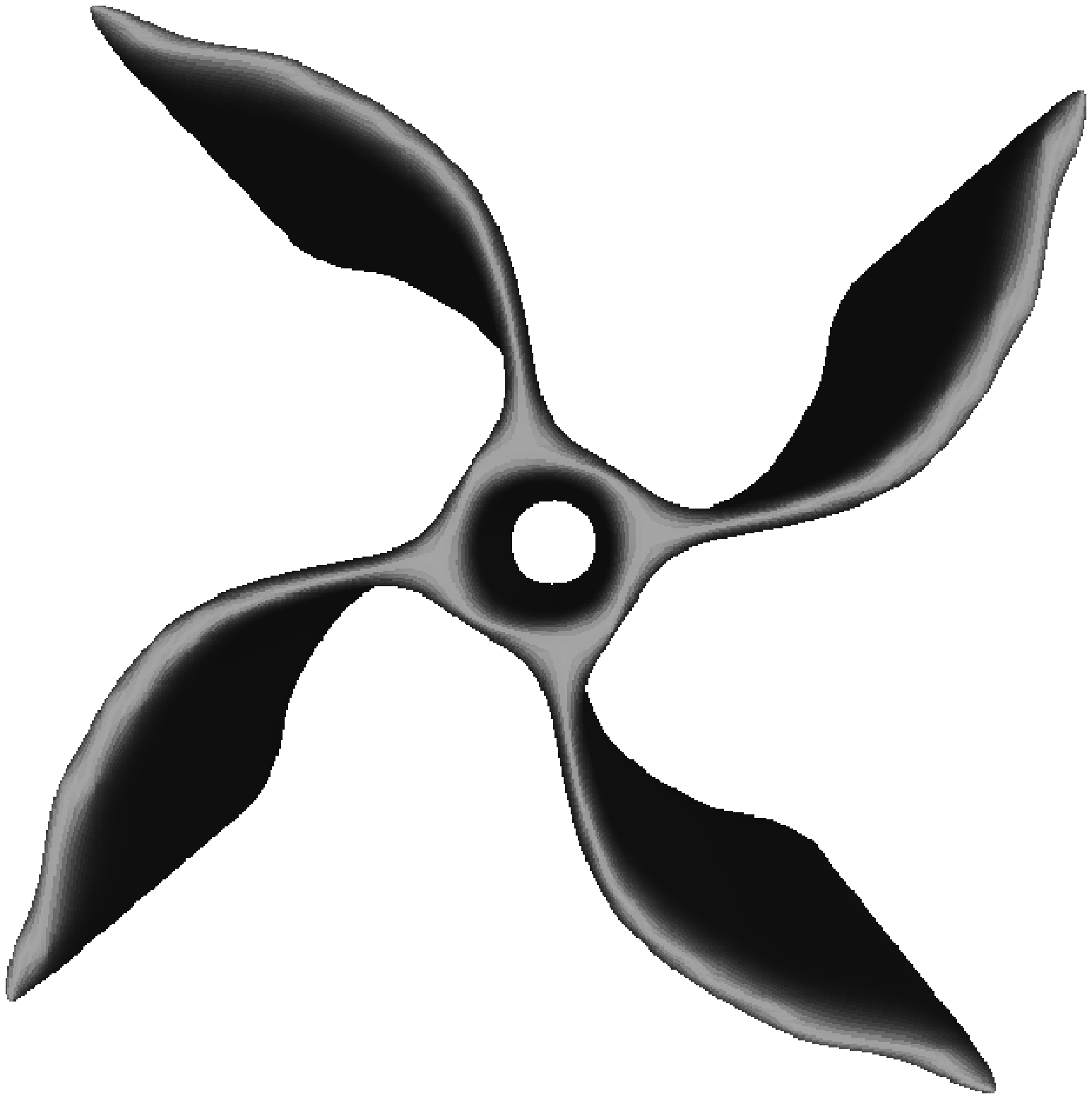,width=\figwidth} } \hspace{0.2in}
\subfigure{\psfig{file=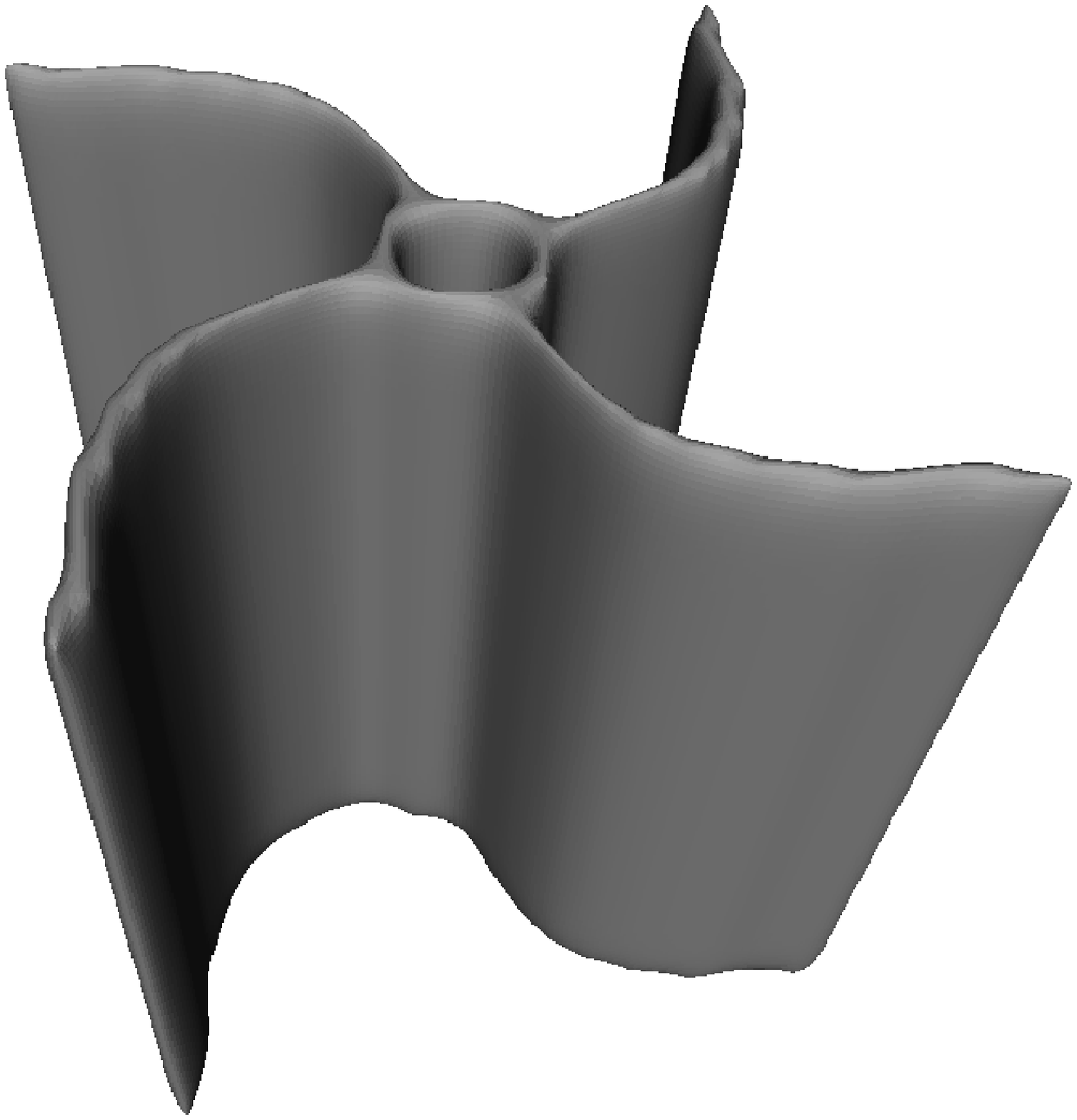,width=\figwidth} }}
\caption{Example with 50 Laplacian smoothing steps applied.} \label{fig:target-smoothed}
\mbox{\subfigure{\psfig{file=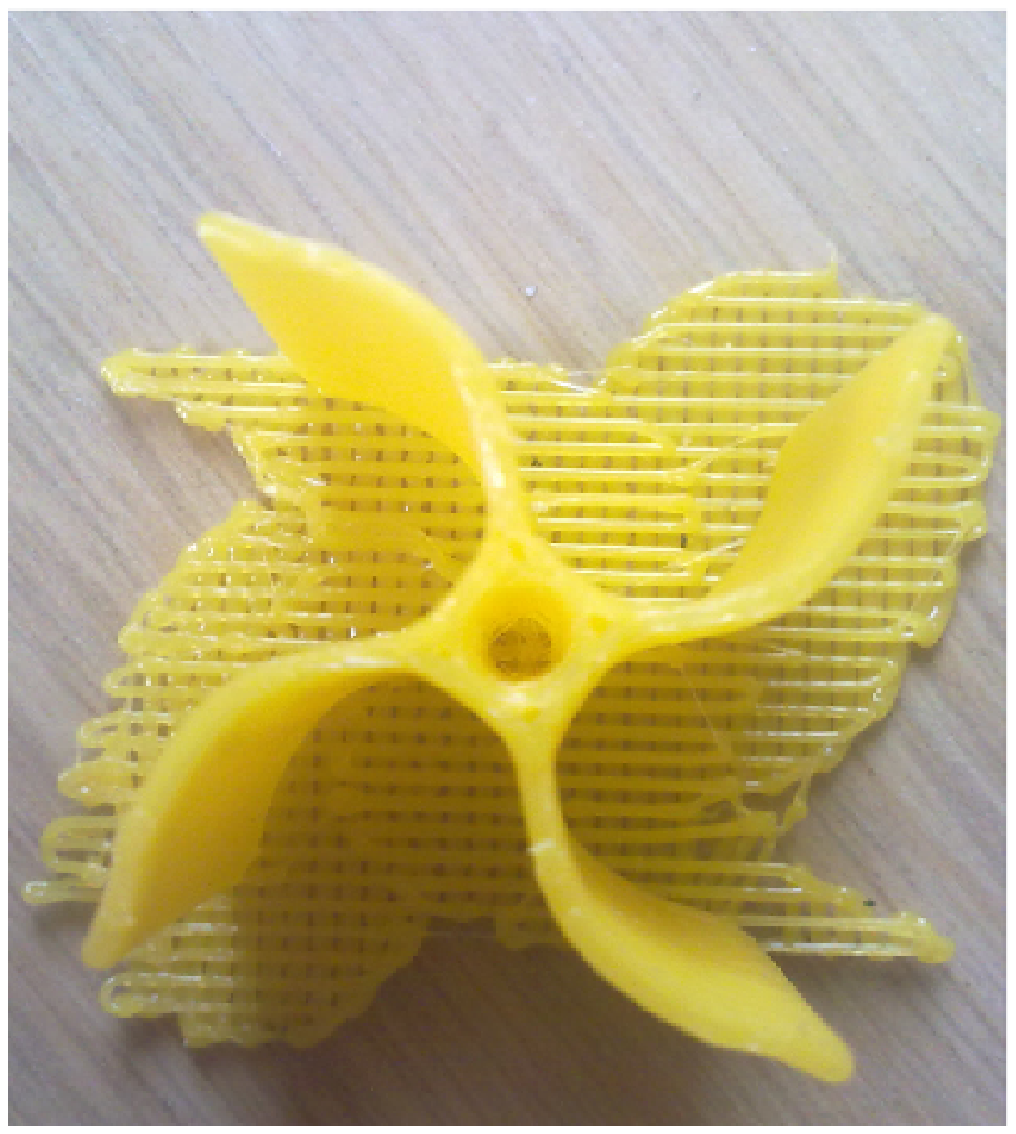,width=1.6in} } \hspace{-0.1in}
\subfigure{\psfig{file=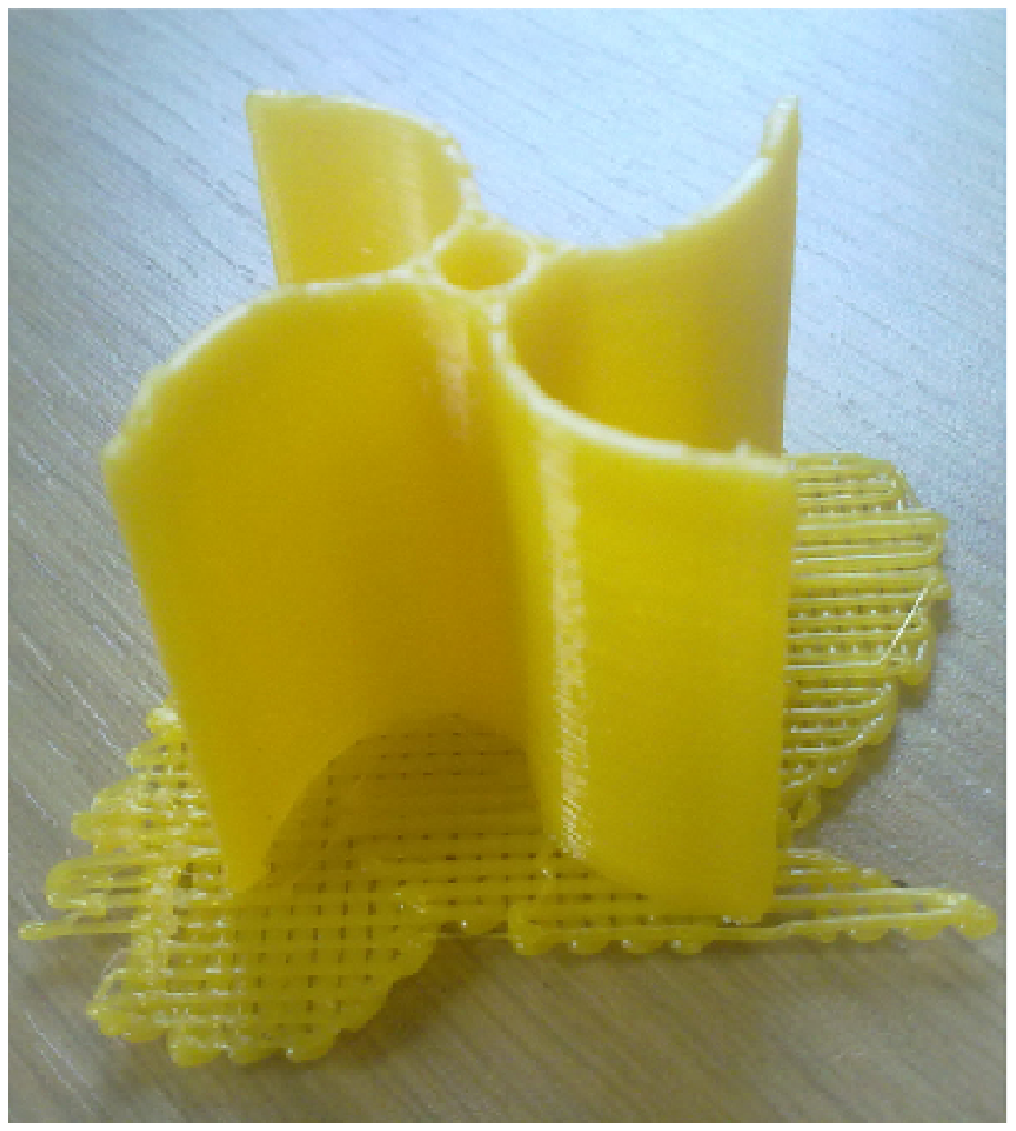,width=1.6in} }}
\caption{Example smoothed and printed by a three-dimensional printer; $30\times30\times30mm$; $27mins$ printing time.} \label{fig:target-printed}
\end{figure}

Figure~\ref{fig:target-unsmoothed} shows an example phenotype. Figure~\ref{fig:target-smoothed} shows the same phenotype with 50 Laplacian smoothing steps subsequently applied to the object with MeshLab\footnote{MeshLab is an open source, portable, and extensible system for the processing and editing of unstructured 3D triangular meshes. http://meshlab.sourceforge.net}. Figure~\ref{fig:target-printed} shows the smoothed object after fabrication by a three-dimensional printer.  

The genetic algorithm (GA) used herein proceeds with a population of 20 individuals, a maximum mutation step size of $\pm10$, and a per allele mutation rate of 25\%. A tournament size of 2 is used for both selection and replacement.

Following previous work on constructing surrogate models \cite{Bull:1999}, here a 3 layer MLP-based artificial neural network is used; composed of 10 input neurons, 5 hidden neurons, 1 output neuron, and trained with backpropagation. The model input is the genome (scaled [-1,1]) and the computed output is the predicted fitness. Initially the entire population is evaluated on the real fitness function and the model is trained using backpropagation for 1,000 epochs; where an epoch consists of randomly selecting, without replacement, an individual from the evaluation set and updating the model weights. Each generation thereafter, the fittest unevaluated individual and a randomly chosen unevaluated individual are evaluated on the real fitness function and the model is iteratively retrained from the entire set of evaluated [genotype, real-fitness] pairs. The model parameters, $\beta=0.3$, $\theta=0$, $elasticity=1$, $calming~rate=1$, $momentum=0$, $elasticity~rate=0$.
\section{Experimentation}
\subsection{Target-Based Evolution}
\begin{figure}[t]
\centering 
\psfig{file=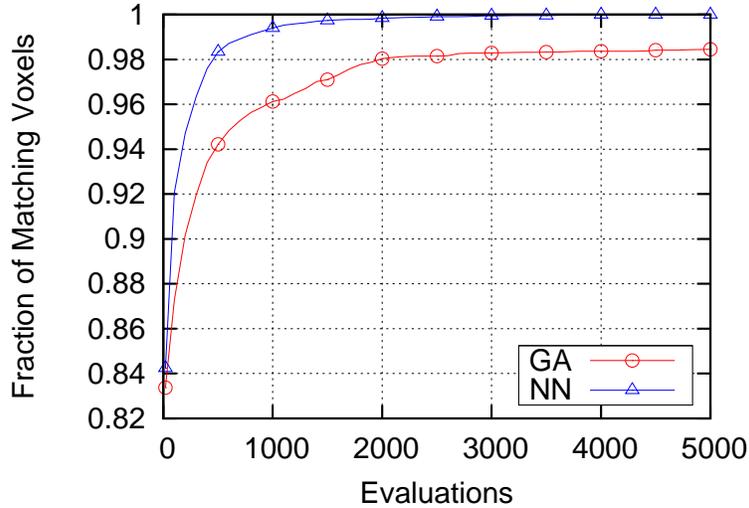,width=\graphwidth}
\caption{Target-based evolution. Fittest GA (circle) and NN surrogate model (triangle) treatments.}
\label{fig:target-res}
\end{figure}

We begin by exploring a target-based approach where the fitness of an individual is the fraction of voxels matching the target object in Figure~\ref{fig:target-unsmoothed}. A total of 20 experiments were run and the results are shown in Figure~\ref{fig:target-res}. Similar to \cite{CluneLipson:2011}, a large number of voxels are quickly matched, however here the target object is not identifiable until approximately 99\% are set correctly. As such, the small differences in fitness between the treatments represent substantial differences in whether the target object is recognisable. As can be seen, the number of matching voxels with the surrogate assisted approach (NN) increases with fewer evaluations than the GA-only approach. In addition, the GA-only approach failed to achieve greater than 99\% performance in 5 of the experiments; whereas with the use of the surrogate model, greater than 99\% was achieved in all experiments. The average number of evaluations required by the GA to reach 99\% matching voxels\footnote{For experiments where the GA did not achieve greater than 99\% fitness within 10,000 evaluations, a value of 10,000 is used.} ($M=3735$, $SD=3922$, $N=20$) is significantly greater than the surrogate model approach ($M=770$, $SD=215$, $N=20$) using a two-sample {\em t}-test assuming unequal variances, $t(19)=3.376$, $p\le.0032$, showing that the model is able to identify an exploitable relationship between the genotype and fitness and the GA can use this for faster convergence to the target shape.
\subsection{Tip Speed-Based Evolution}
\begin{figure}[t]
\centering 
\psfig{file=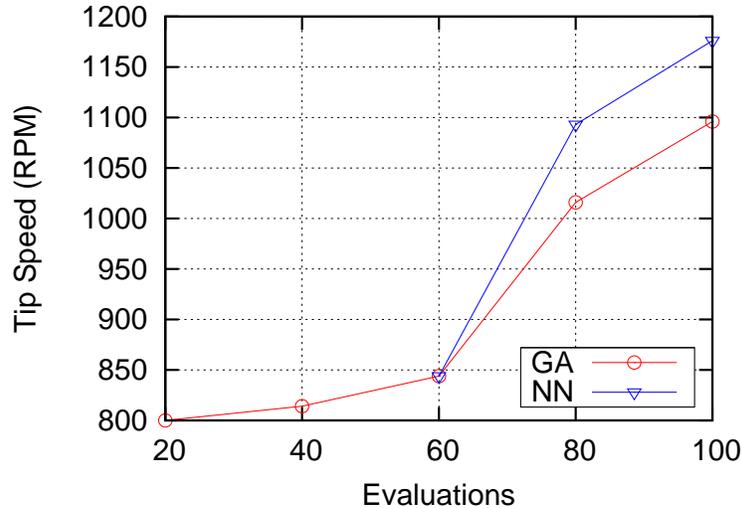,width=\graphwidth}
\caption{Tip Speed-based evolution. Fittest GA (circle) and NN surrogate model (triangle) treatments.}
\label{fig:real-res}
\end{figure}

\begin{figure}[t]
\centering 
\subfigure[1st Gen]{\label{fig:ga-g1}\psfig{file=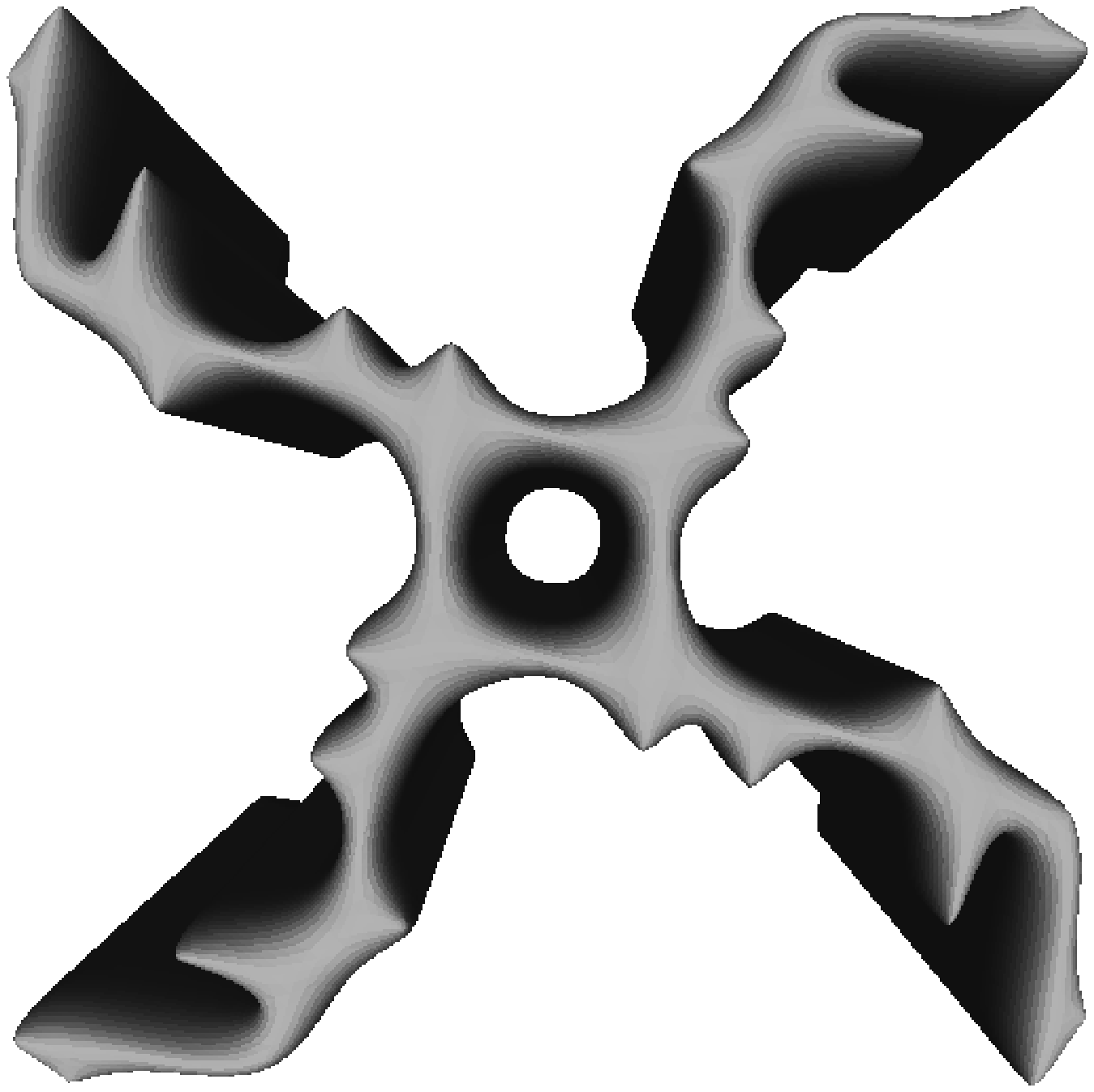,width=\smallfigwidth}}
\subfigure[2nd Gen]{\label{fig:ga-g2}\psfig{file=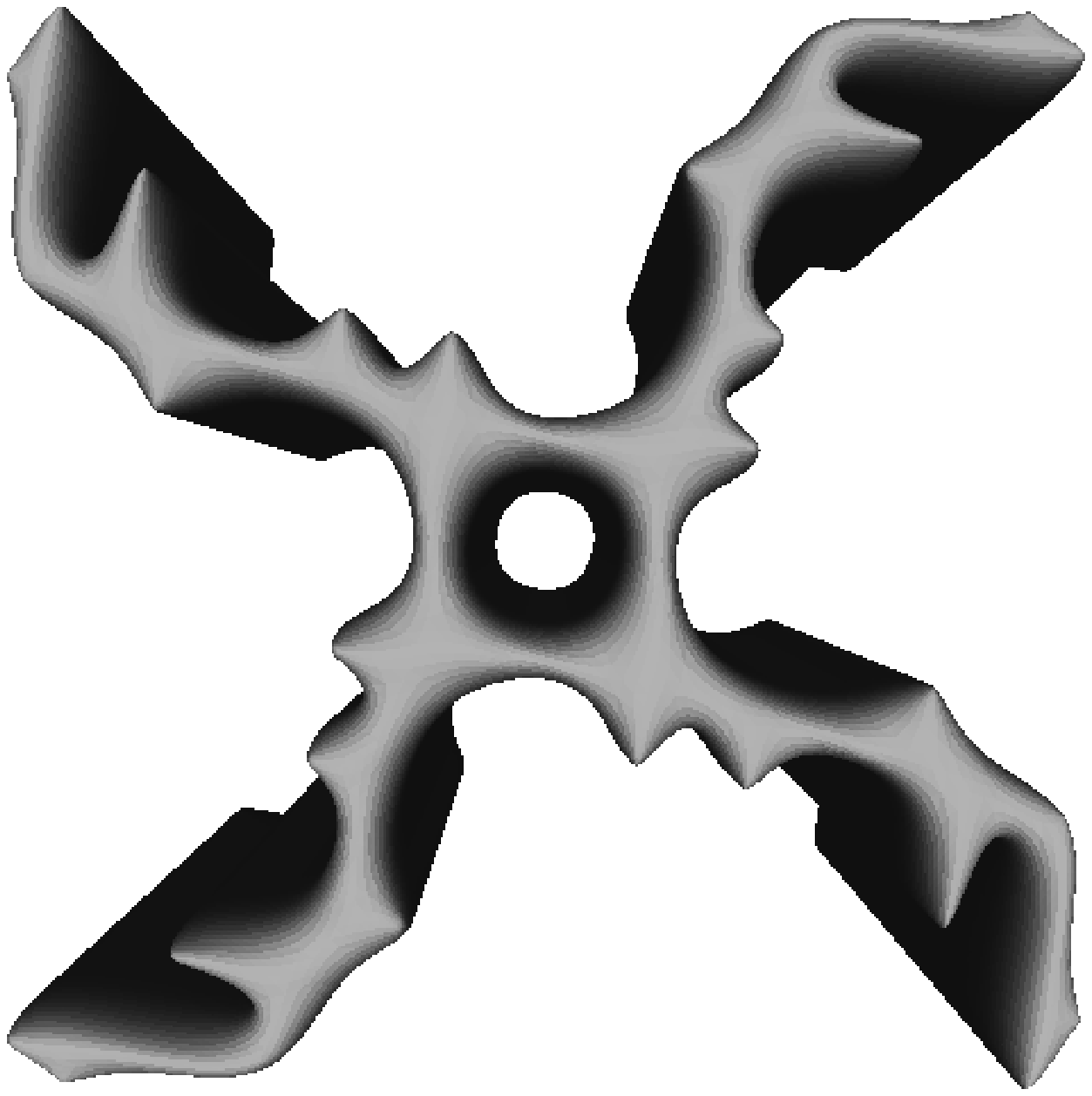,width=\smallfigwidth}}
\subfigure[3rd Gen]{\label{fig:ga-g3}\psfig{file=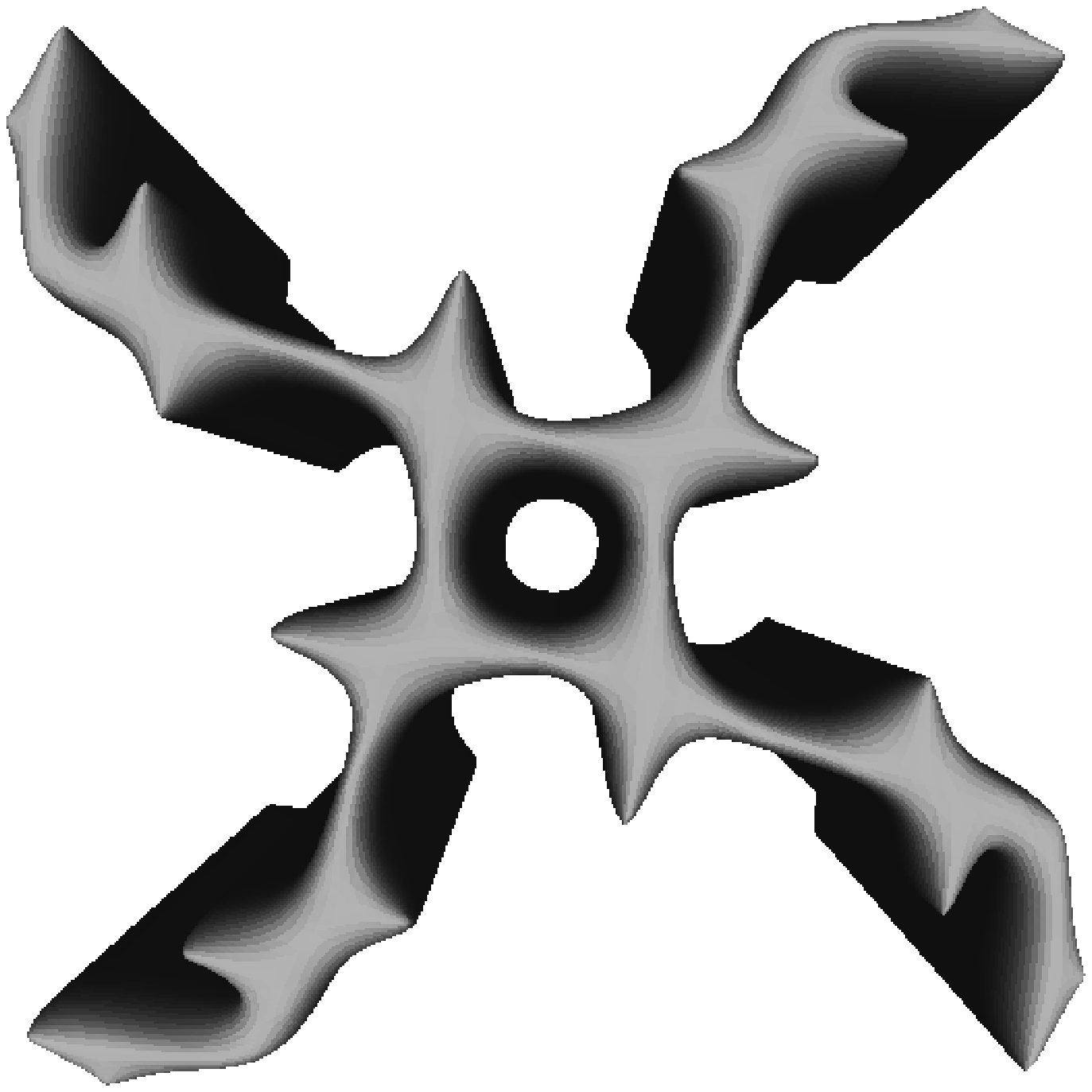,width=\smallfigwidth}}
\subfigure[4th Gen]{\label{fig:ga-g4}\psfig{file=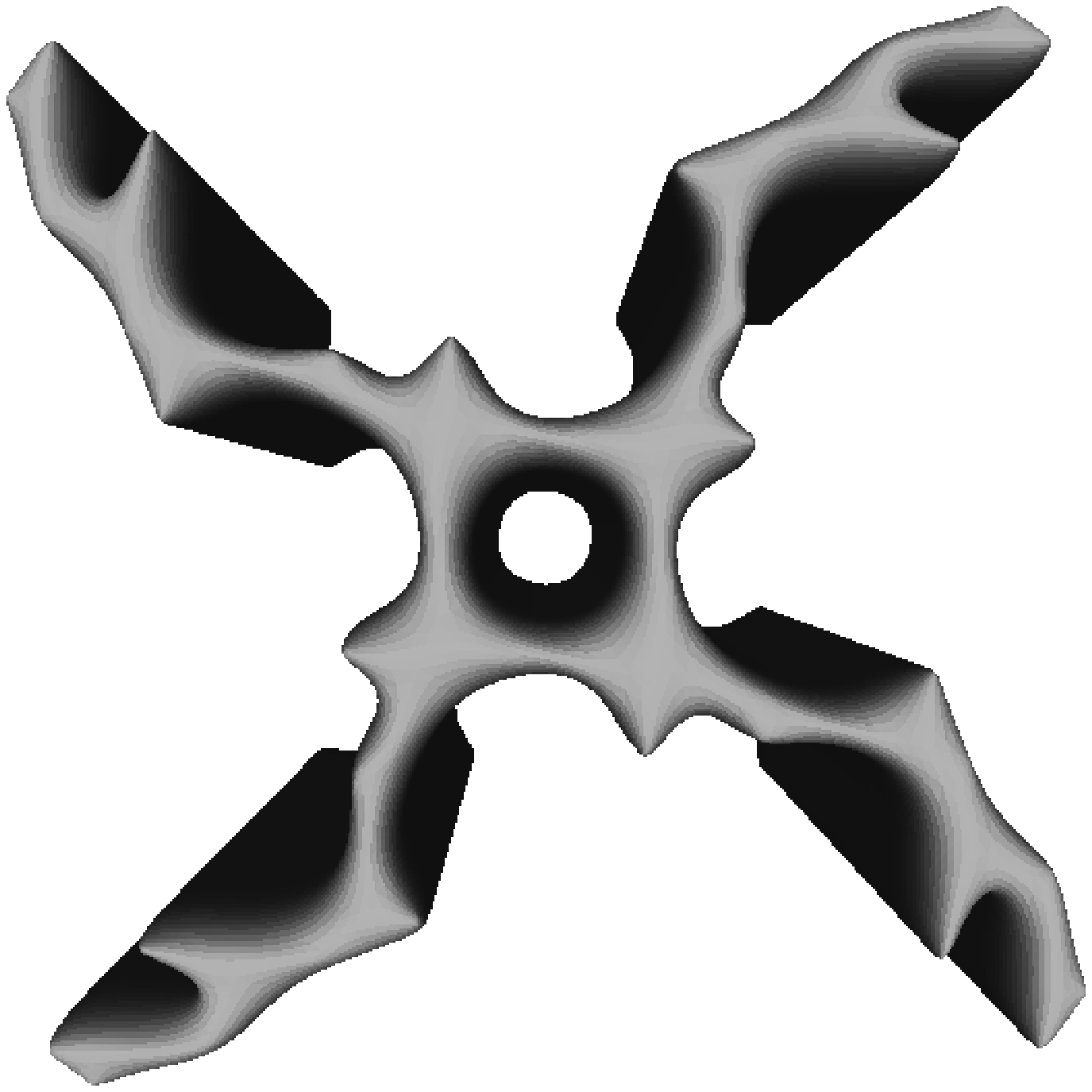,width=\smallfigwidth}}
\subfigure[5th Gen]{\label{fig:ga-g5}\psfig{file=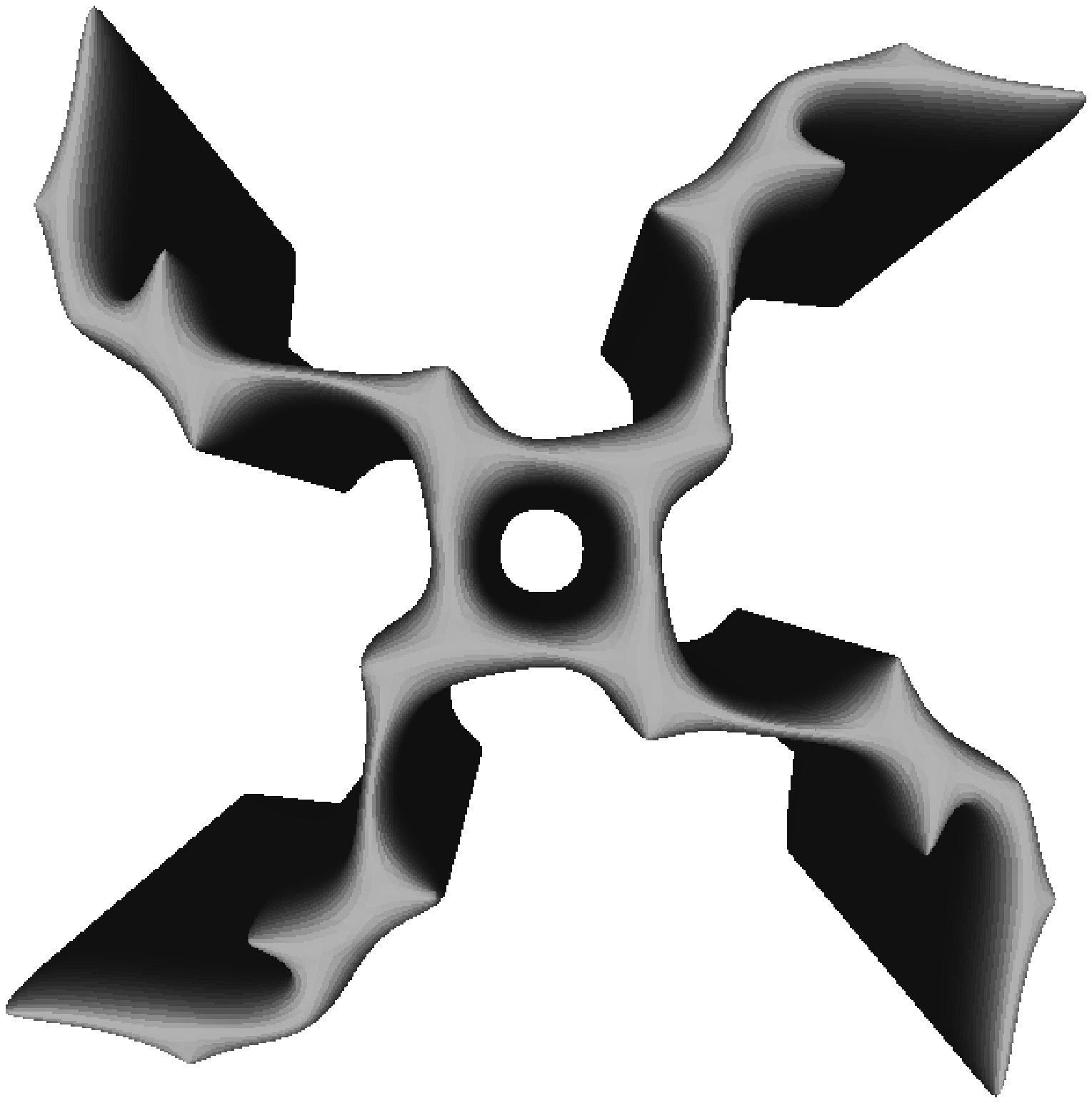,width=\smallfigwidth}}
\caption{The fittest treatments produced by the GA each generation.}
\label{fig:ga-individuals}
\subfigure[4th Gen]{\label{fig:n-g4}\psfig{file=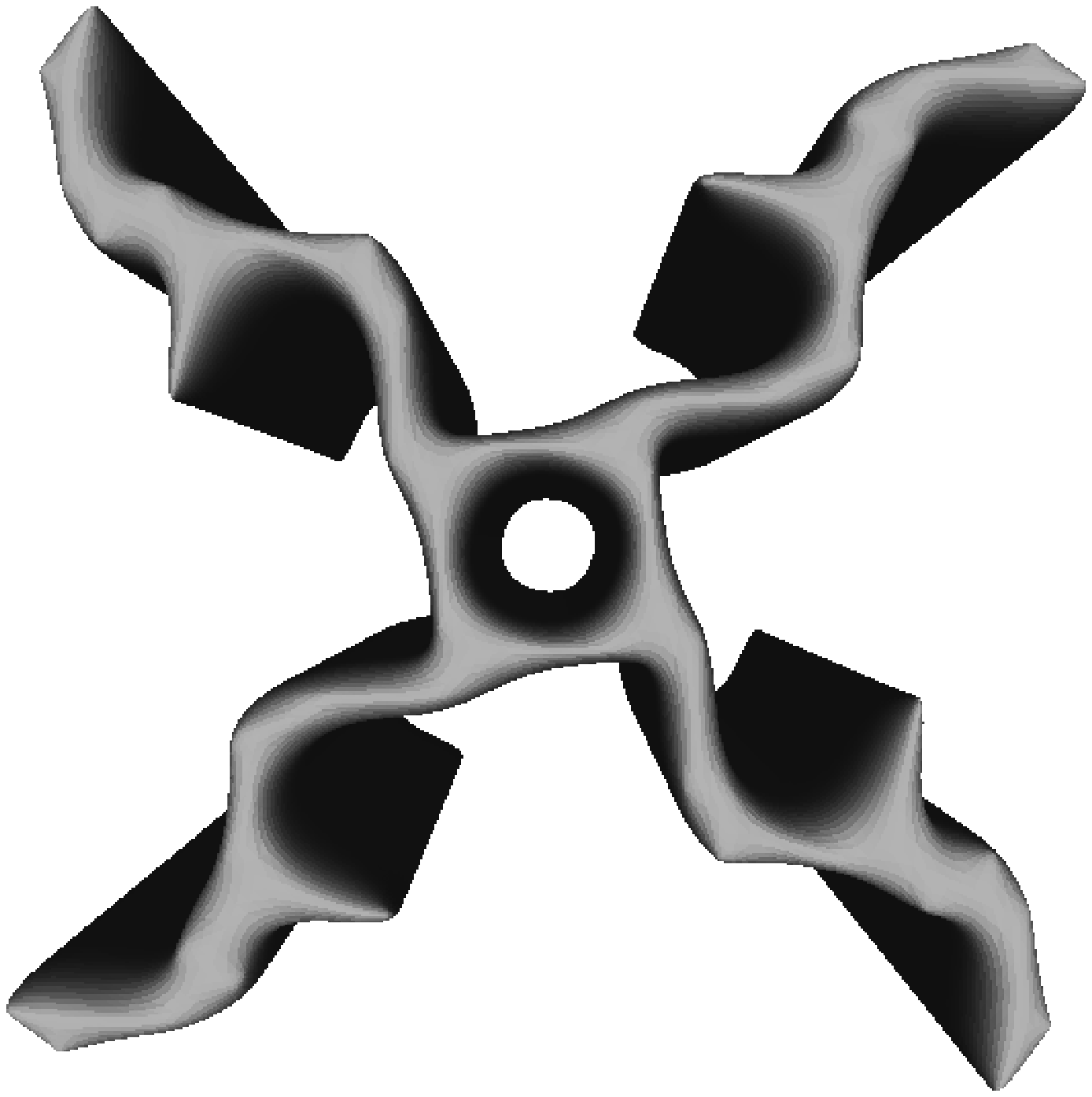,width=\smallfigwidth}}
\subfigure[5th Gen]{\label{fig:n-g5}\psfig{file=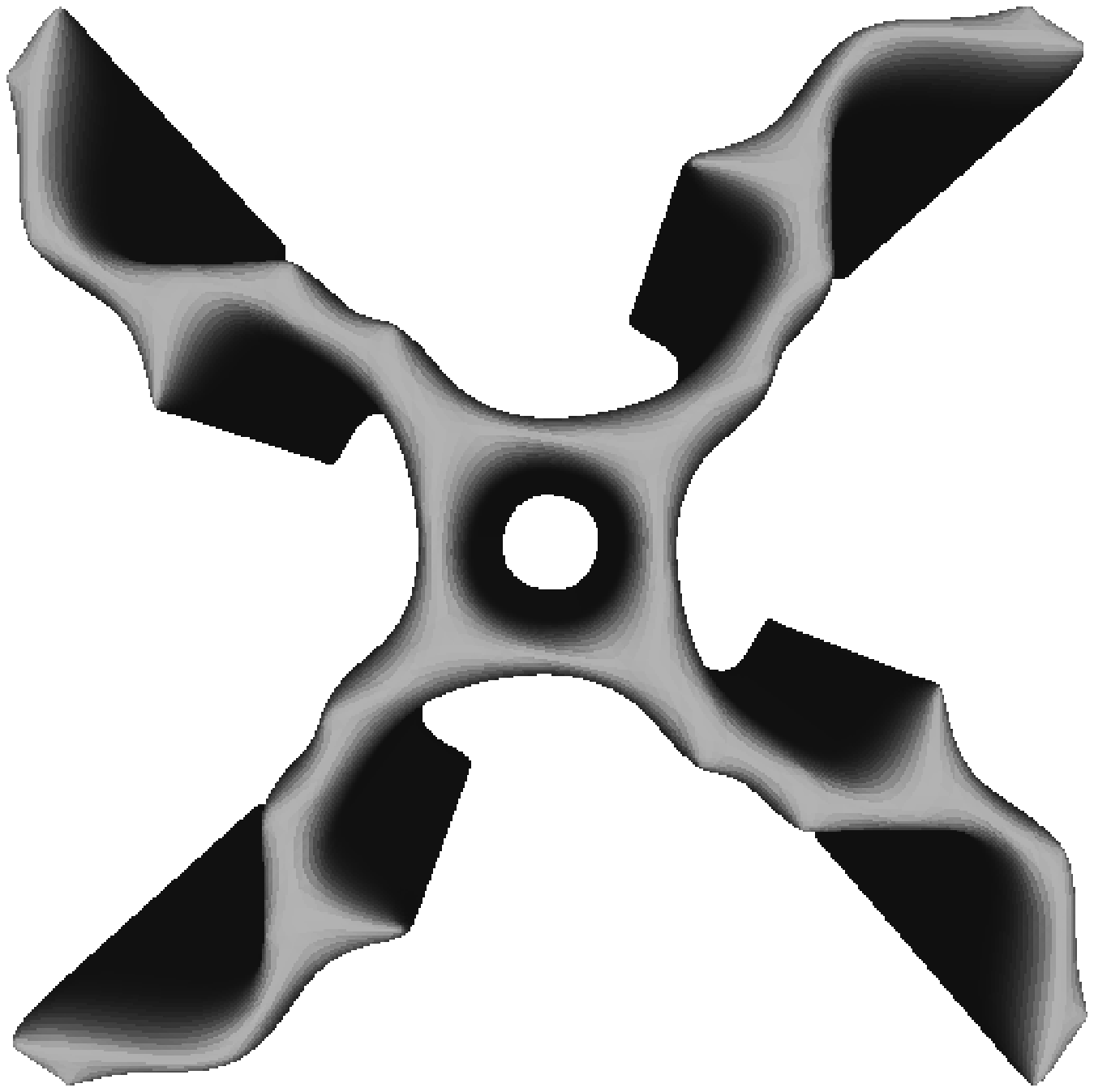,width=\smallfigwidth}}
\caption{The fittest treatments produced by the model each generation.}
\label{fig:model-individuals}
\end{figure}

As a first step towards the evolution of novel VAWTs, here the fitness computation for each individual becomes the maximum tip speed achieved during the application of constant wind generated by an approximated wind tunnel after fabrication by a three-dimensional printer ($300mm$ propeller fan; $3,500rpm$; treatment placed at $30mm$ distance; $4.4m/s$ wind speed). The tip speed is measured in number of revolutions per minute ($rpm$) with a digital photo laser tachometer by placing a $10\times2mm$ strip of reflecting tape on the outer tip of one of the treatment's blades. Initially, 20 random designs are generated, fabricated, and evaluated. Since many of the seed treatments are extremely aerodynamically inefficient (only 2 out of 20 yielded $>0rpm$), the GA is run for 2 further generations before the surrogate model is used for comparison. The initial pilot results from an experiment with the GA and surrogate model are presented in Figure~\ref{fig:real-res}, which shows the maximum tip speed achieved by the fittest treatments in each generation. The GA and model-assisted approach identify increasingly efficient aerodynamic designs, and the surrogate model shows improved performance similar to the prior target-based experiments ($1176rpm$ vs. $1096rpm$ after 100 evaluations). The fittest treatments produced by the GA and surrogate model each generation are shown in Figures~\ref{fig:ga-individuals} and \ref{fig:model-individuals}, respectively.

\begin{figure}[t]
\centering 
\psfig{file=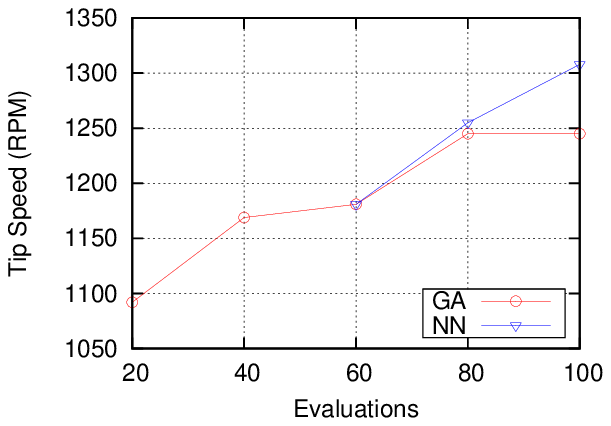,width=\graphwidth}
\caption{Tip Speed-based evolution with $z$-variability. Fittest GA (circle) and NN surrogate model (triangle) treatments.}
\label{fig:zreal-res}
\end{figure}

\begin{figure}[t]
\centering 
\subfigure[1st Gen]{\label{fig:zga-g1}\psfig{file=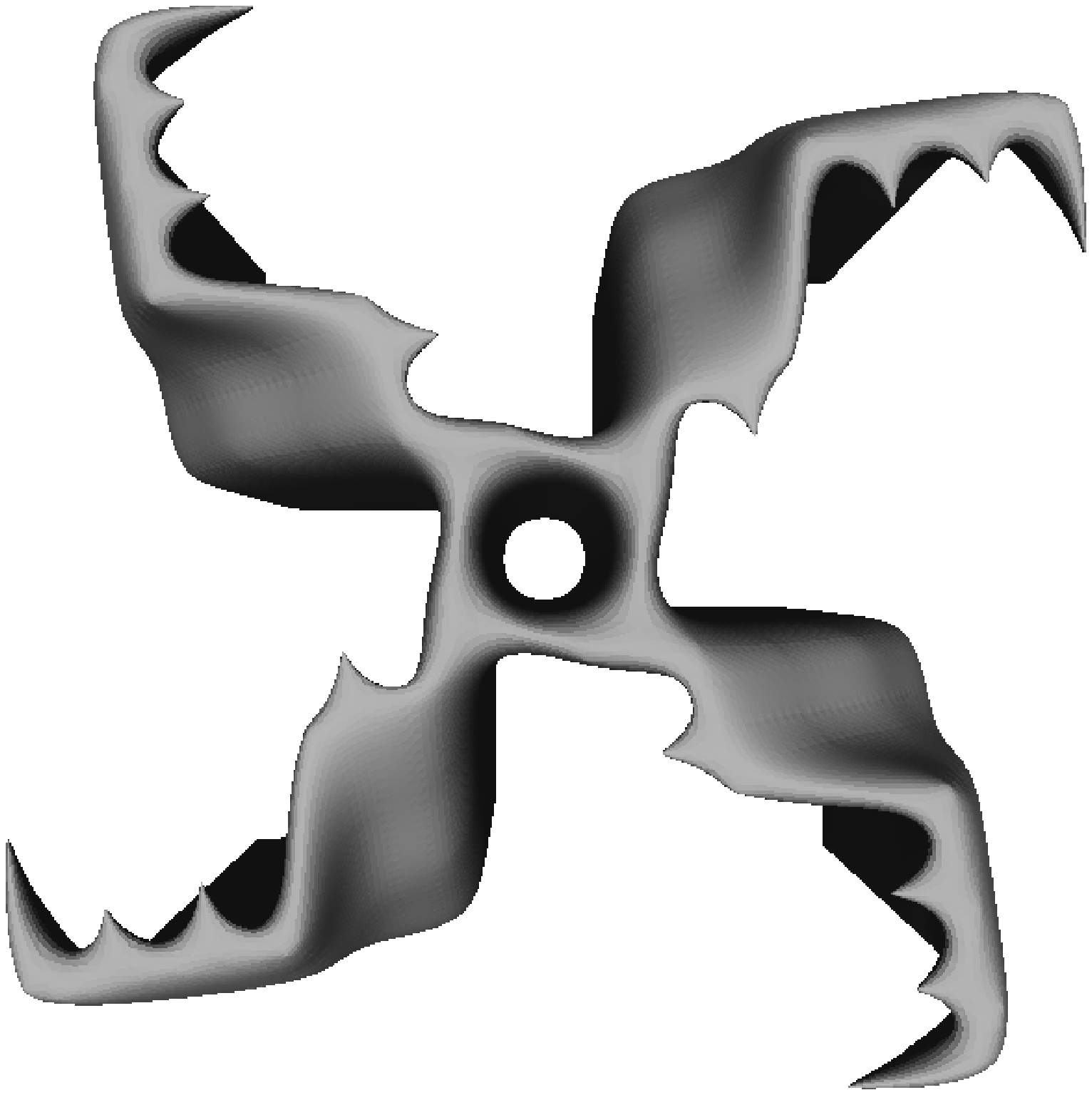,width=\zsmallfigwidth} \psfig{file=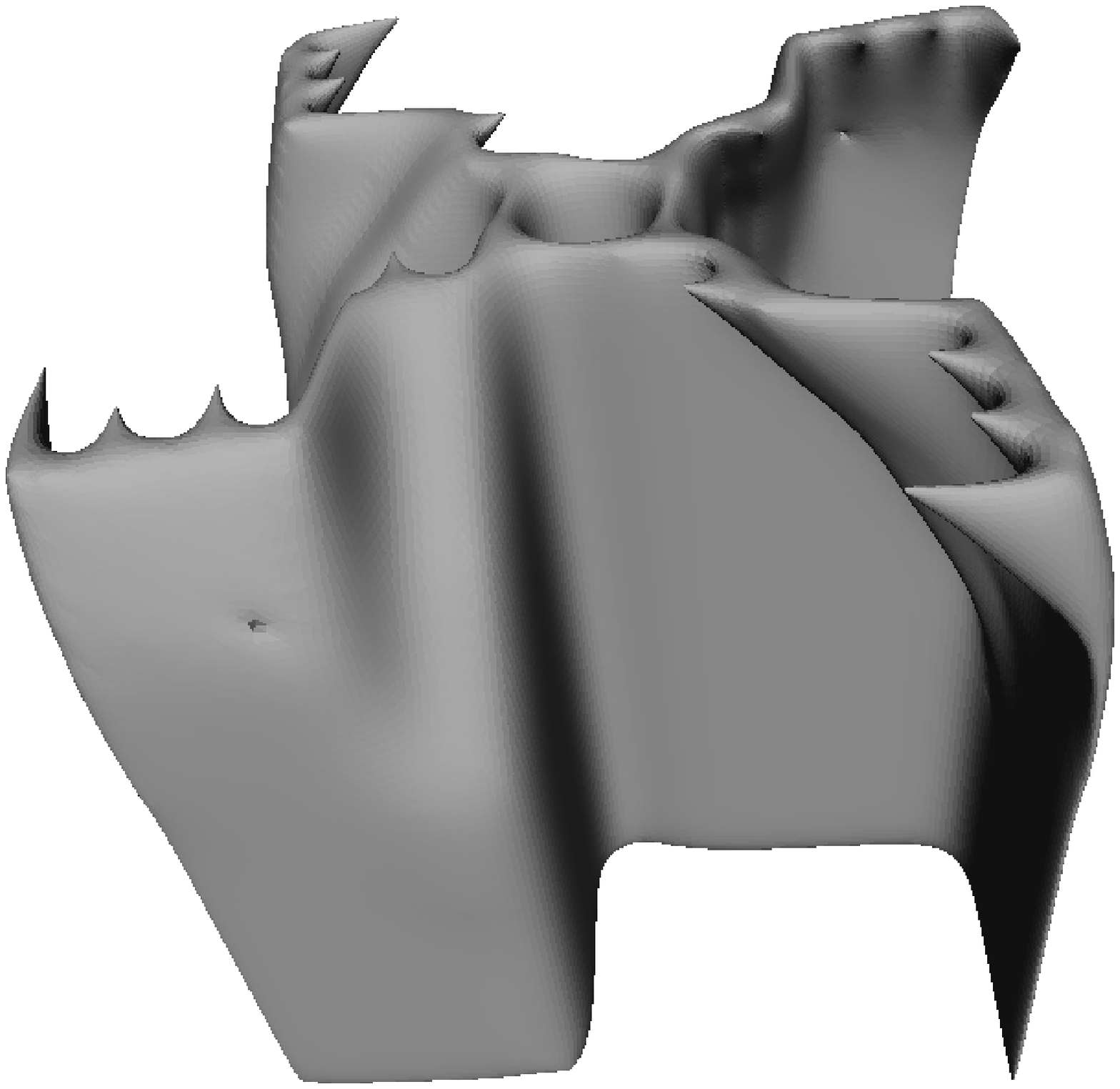,width=\zsmallfigwidth}}
\subfigure[2nd Gen]{\label{fig:zga-g2}\psfig{file=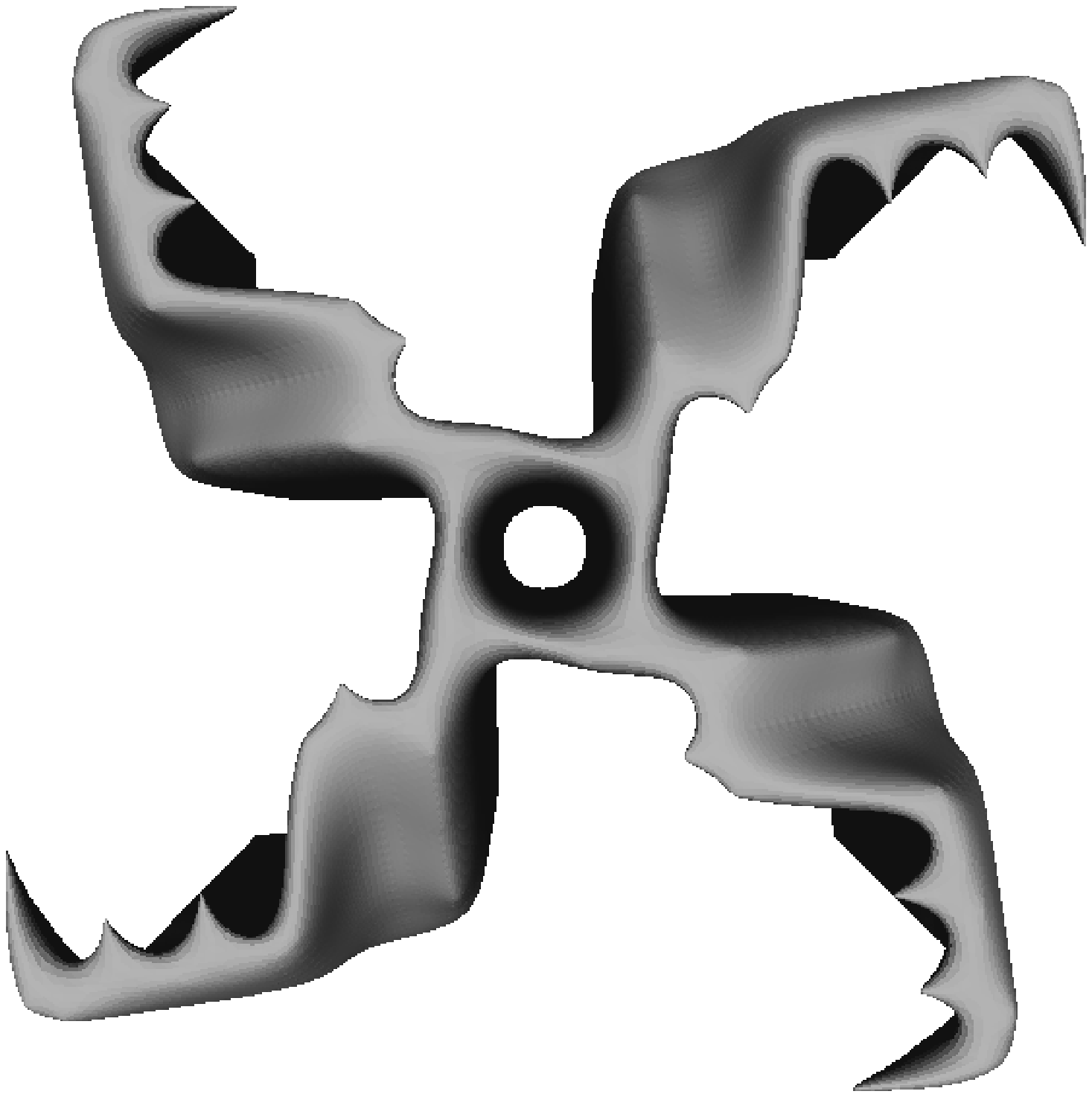,width=\zsmallfigwidth} \psfig{file=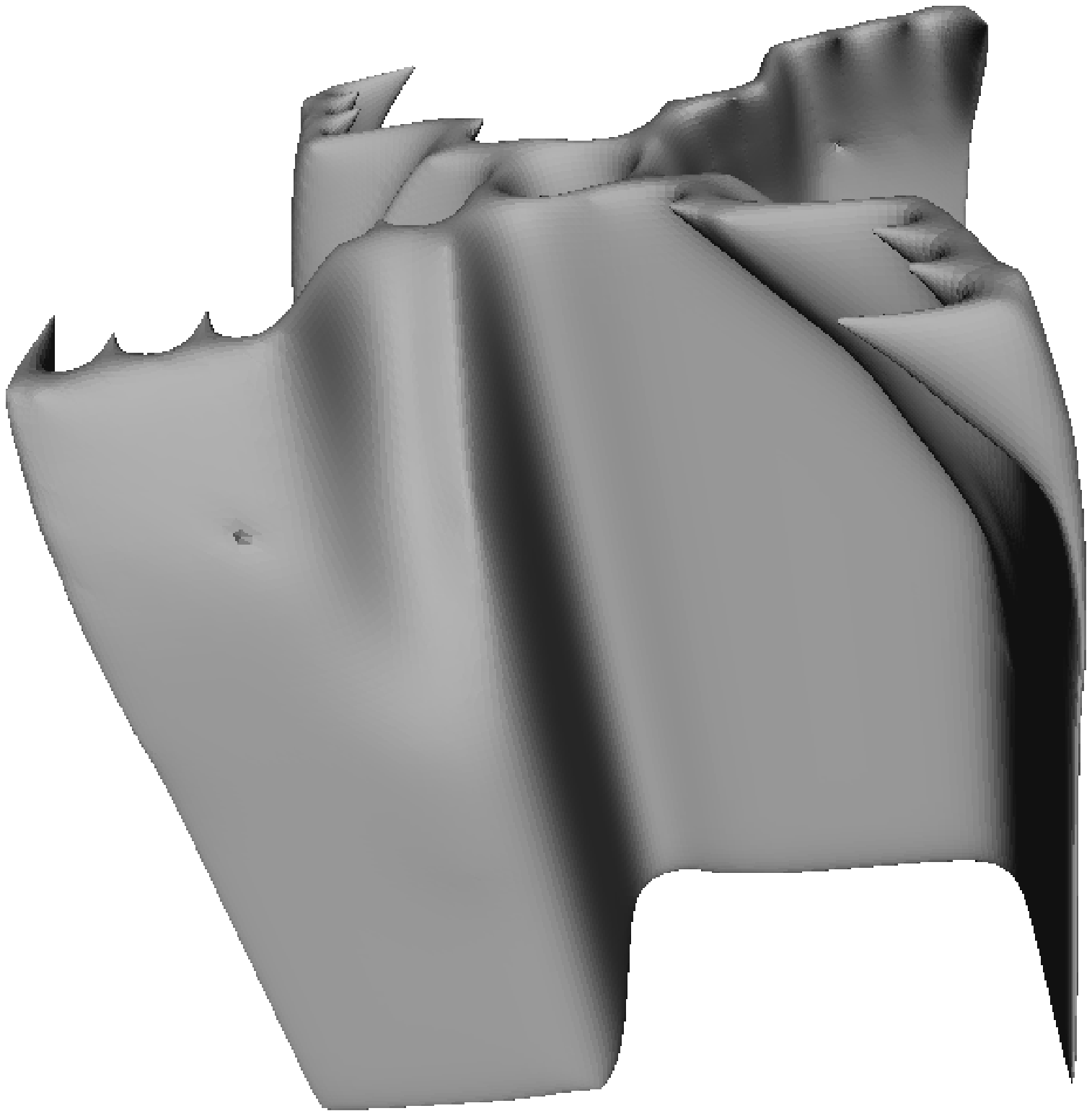,width=\zsmallfigwidth}}
\subfigure[3rd Gen]{\label{fig:zga-g3}\psfig{file=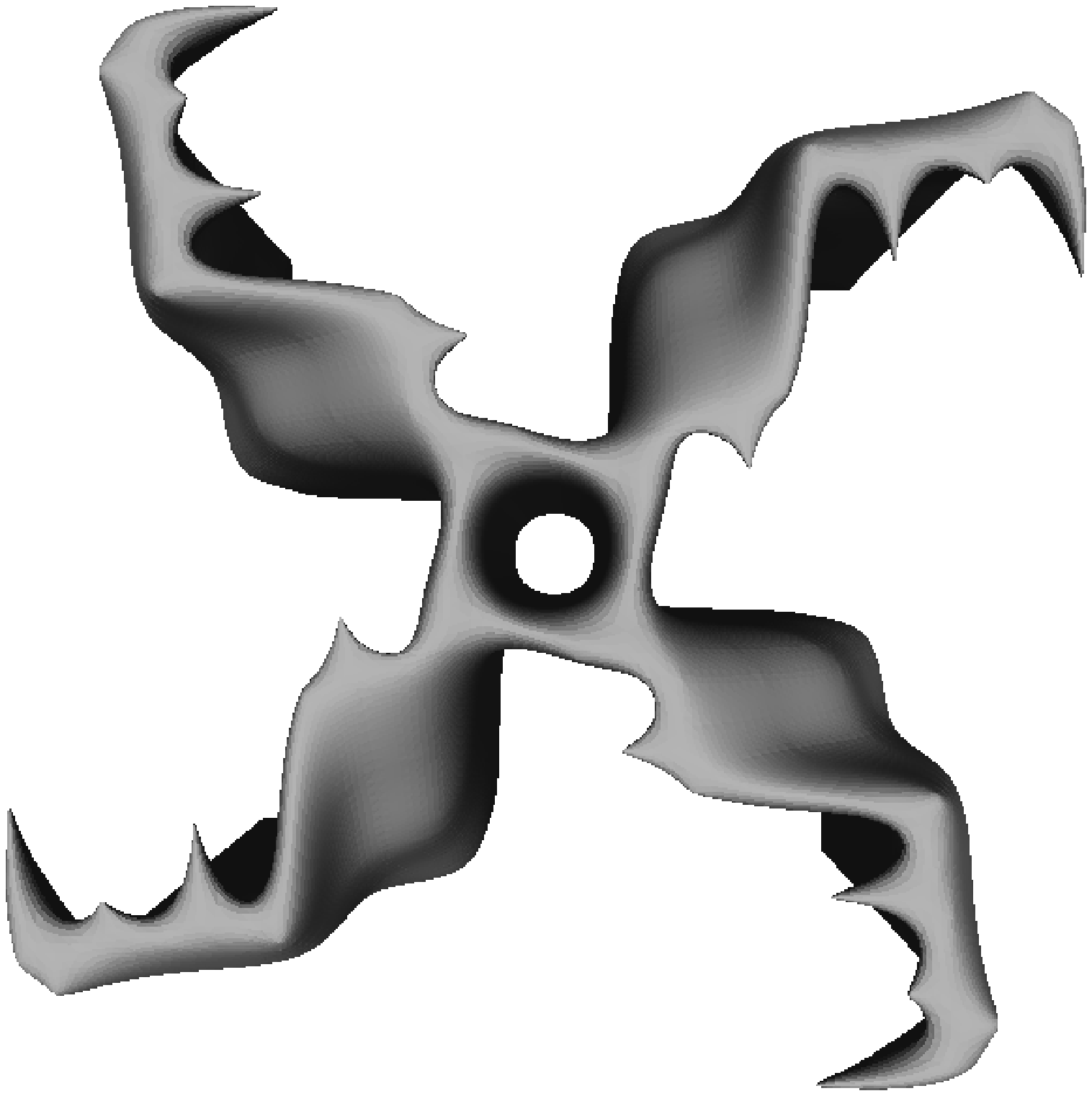,width=\zsmallfigwidth} \psfig{file=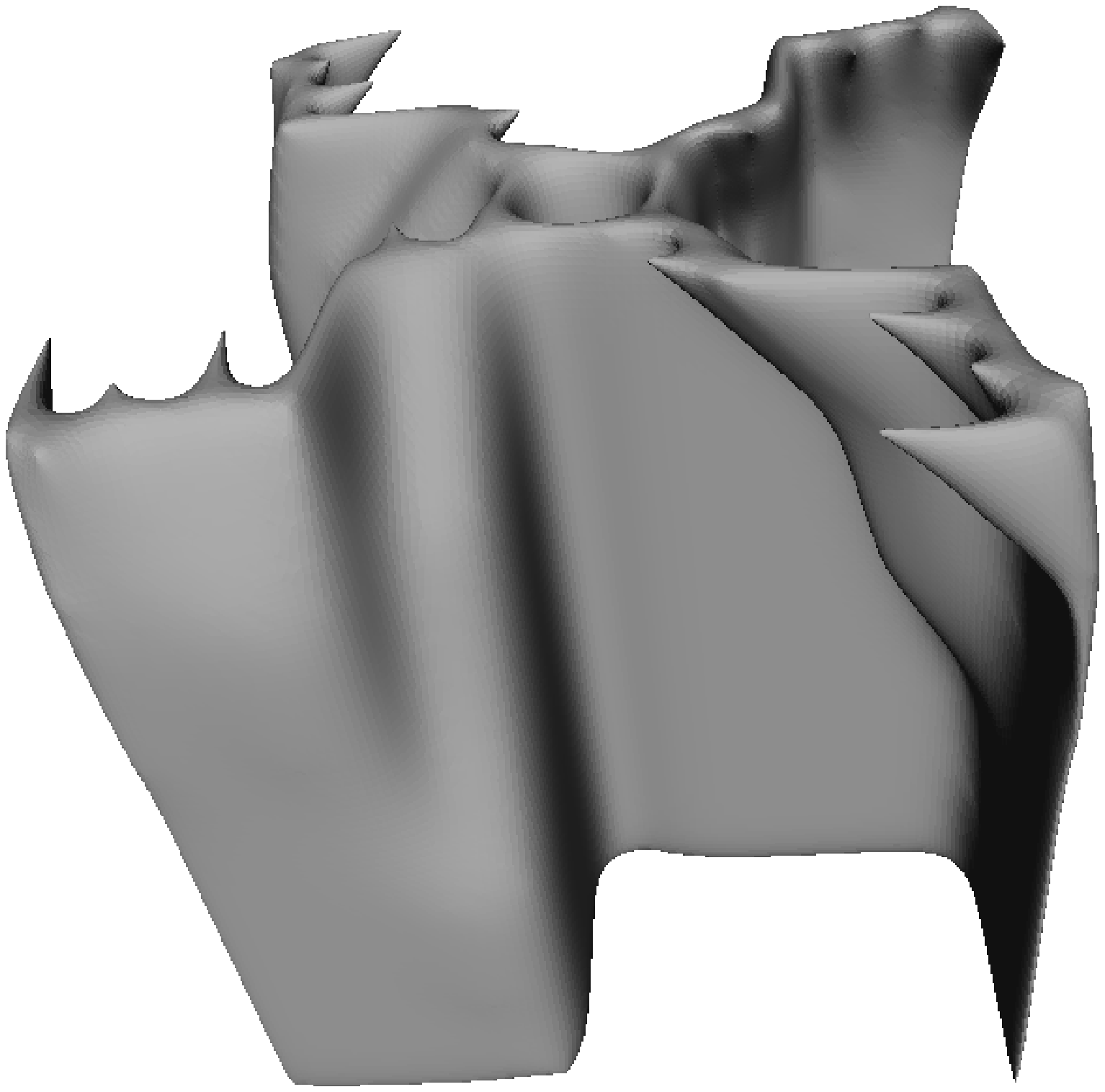,width=\zsmallfigwidth}}
\subfigure[4th/5th Gen]{\label{fig:zga-g45}\psfig{file=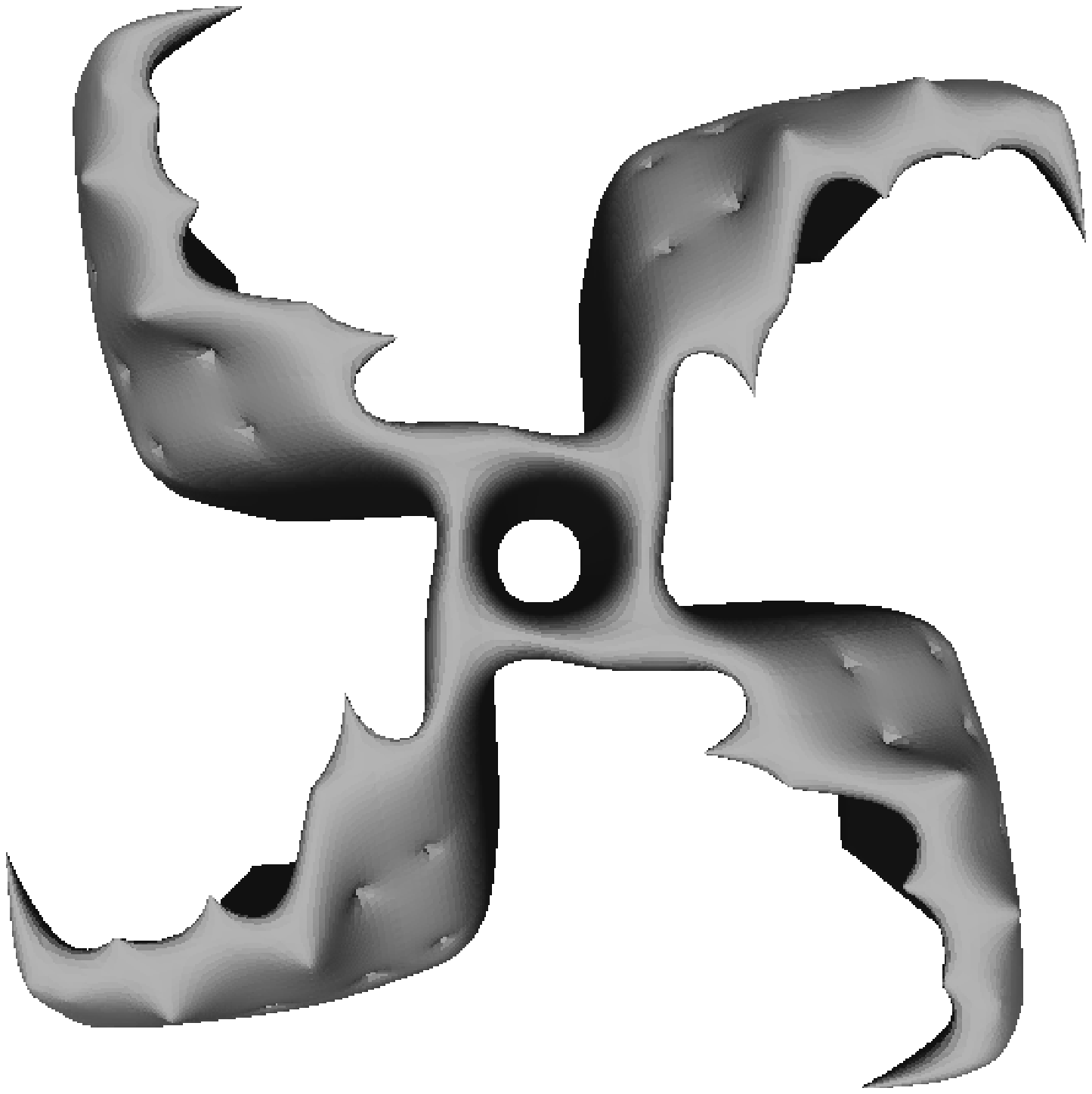,width=\zsmallfigwidth} \psfig{file=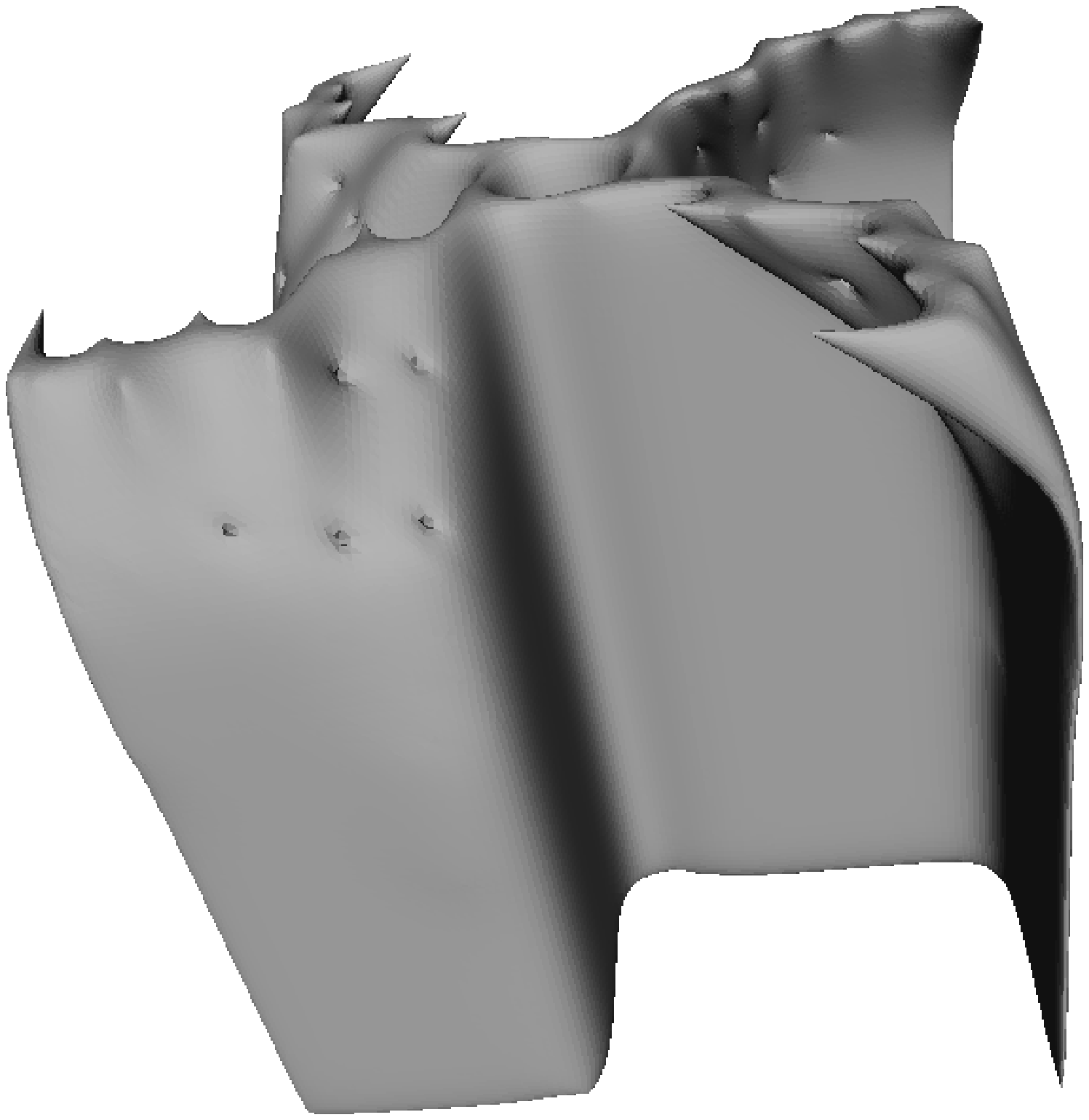,width=\zsmallfigwidth}}
\caption{The fittest treatments with $z$-variability produced by the GA each generation.}
\label{fig:zga-individuals}
\end{figure}

\begin{figure}[t]
\centering 
\subfigure[4th Gen]{\label{fig:zn-g4-top}\psfig{file=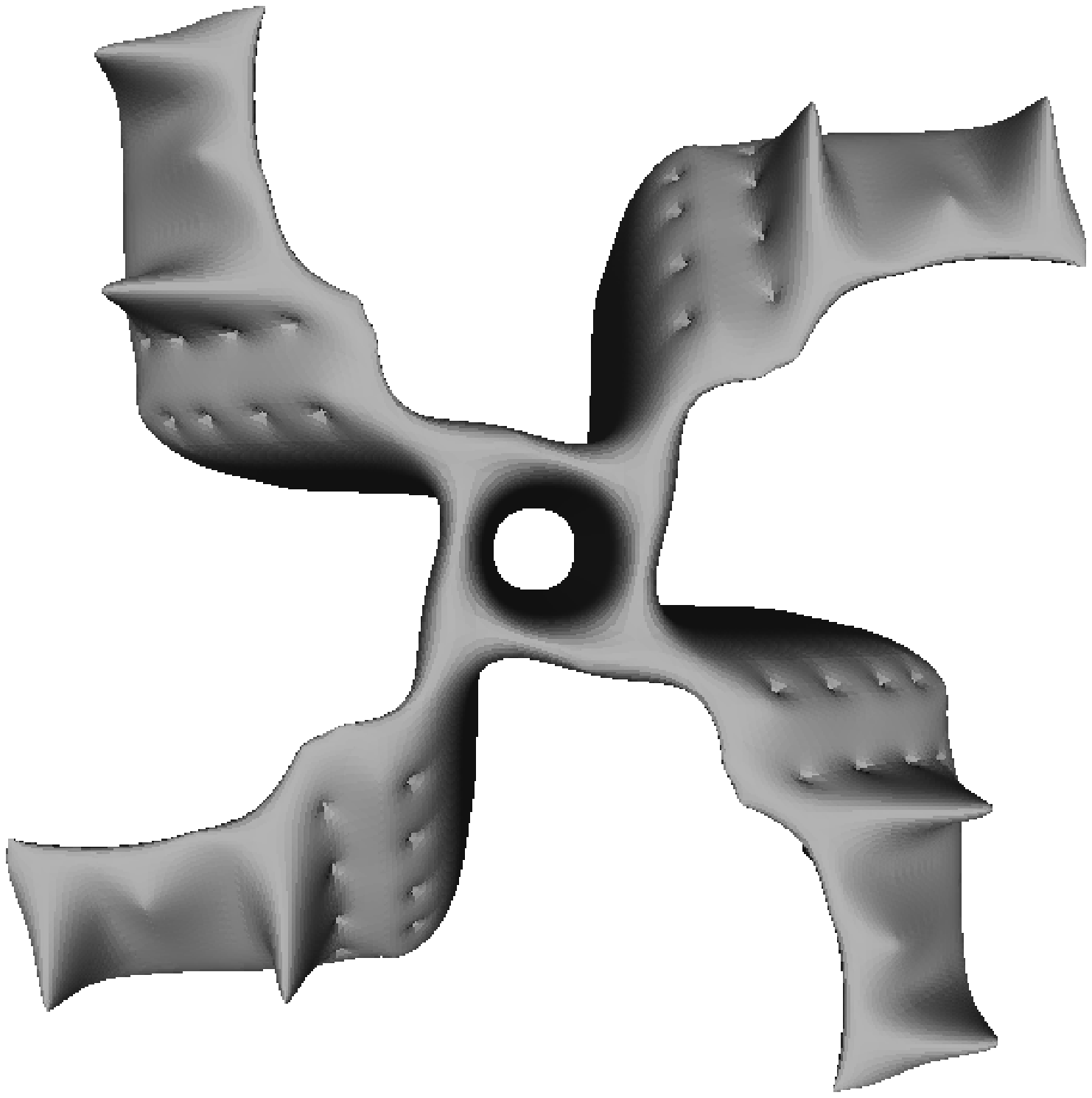,width=\zsmallfigwidth} \psfig{file=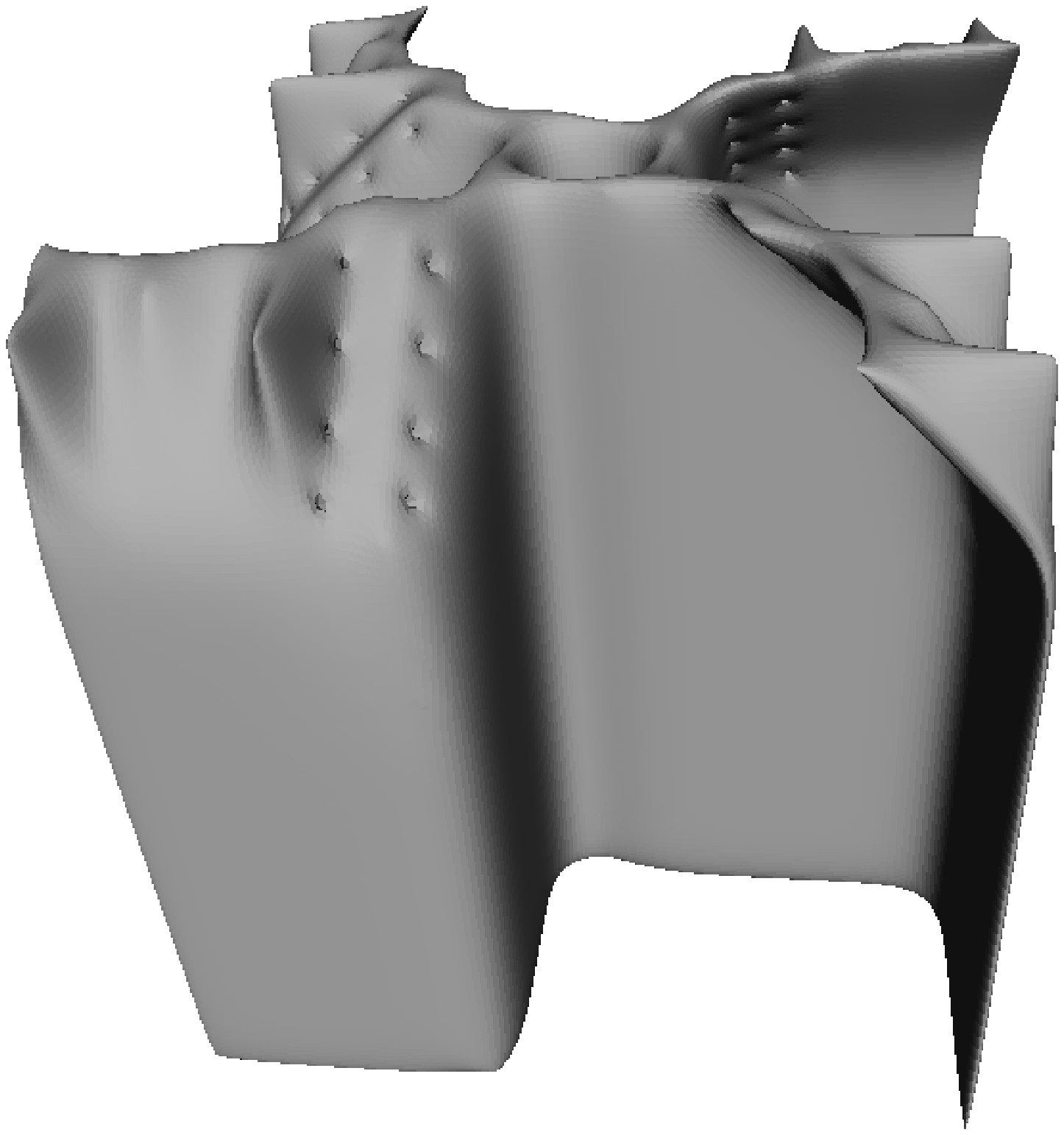,width=\zsmallfigwidth}}
\subfigure[5th Gen]{\label{fig:zn-g5-top}\psfig{file=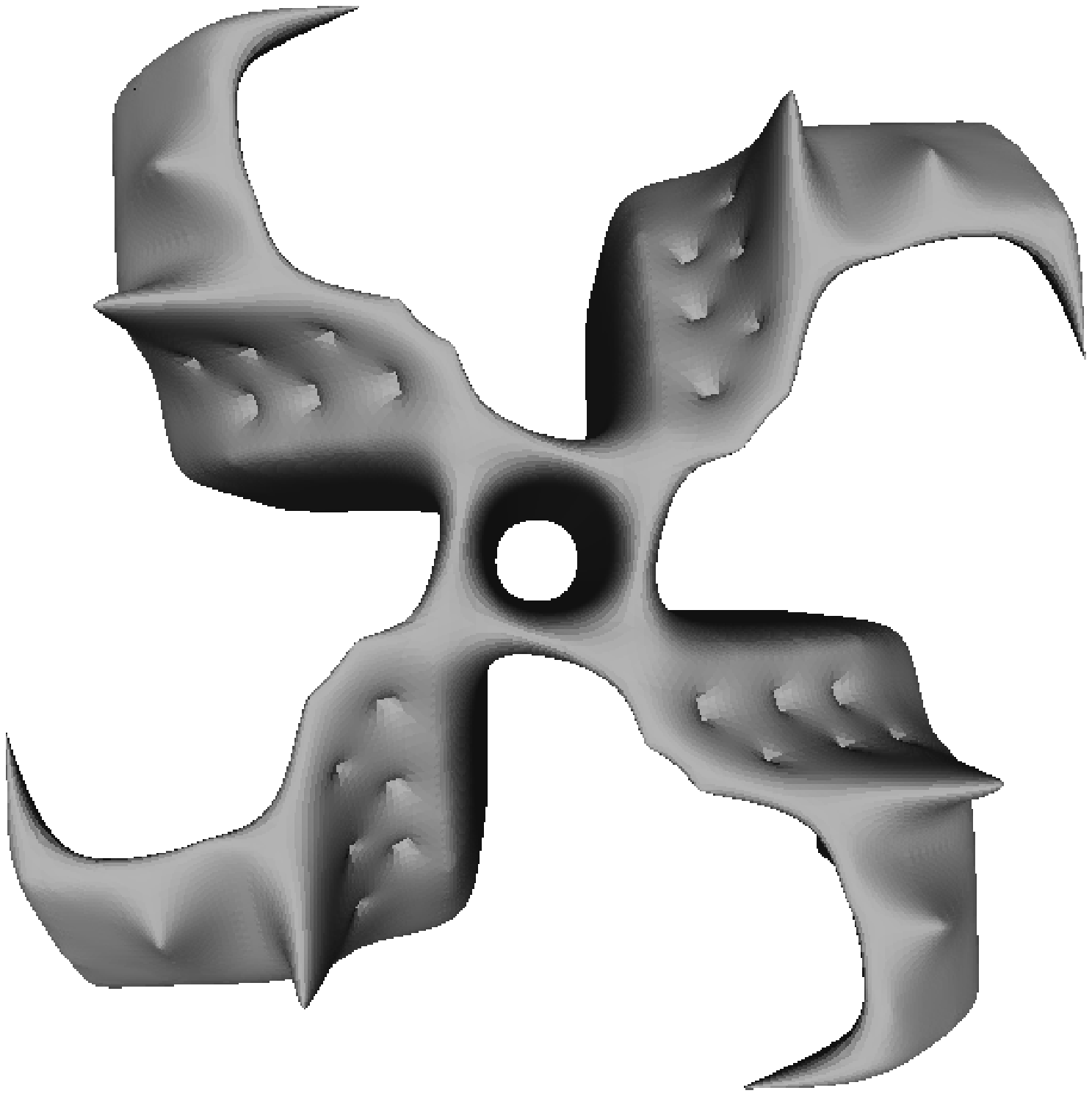,width=\zsmallfigwidth} \psfig{file=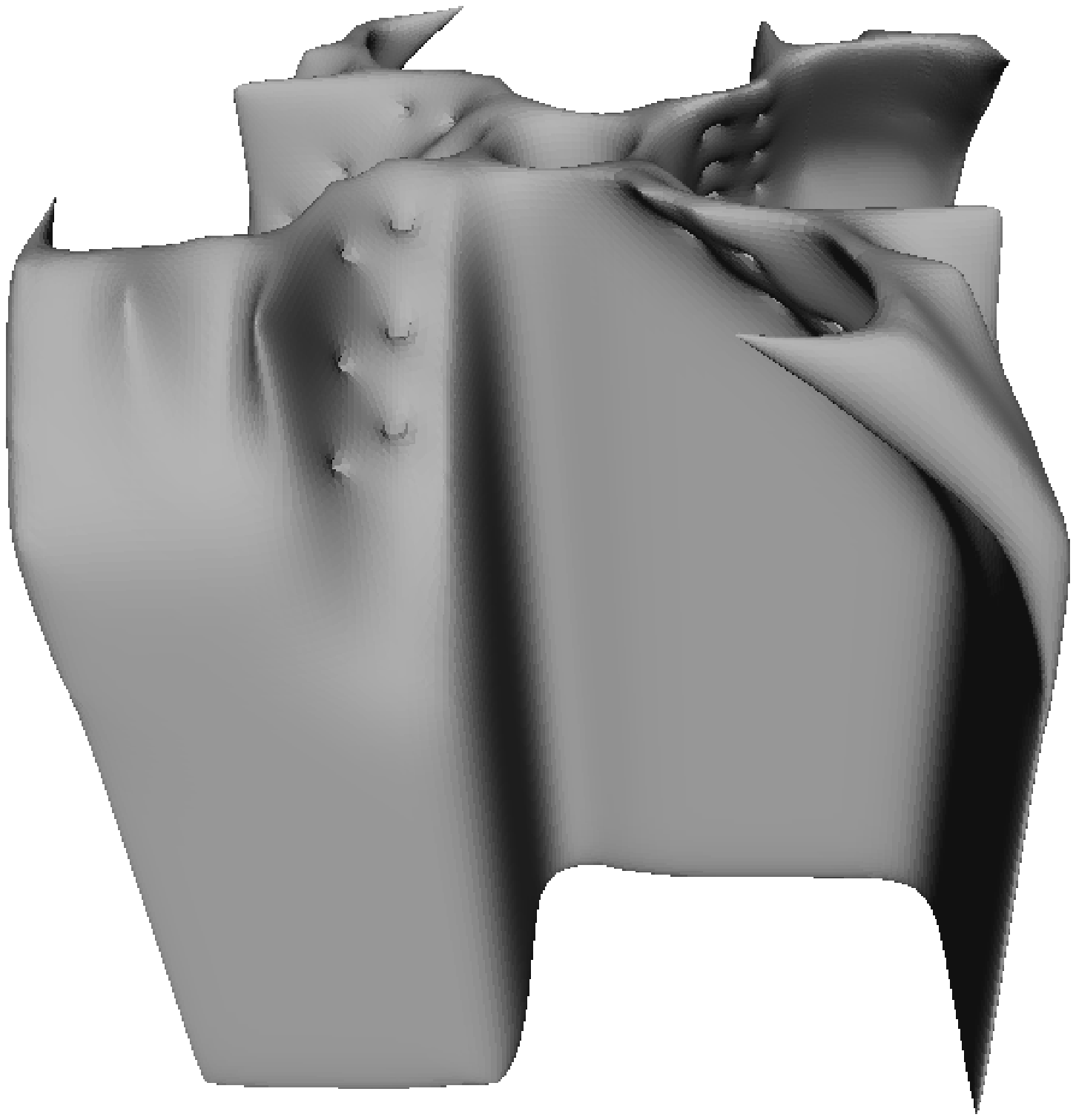,width=\zsmallfigwidth}}
\caption{The fittest treatments with $z$-variability produced by the model each generation.}
\label{fig:zmodel-individuals}
\end{figure}

In order to provide an encoding simple enough for the surrogate model to map over, the turbine representation used so far has restricted the morphology in the $z$ dimension. However, more flexibility potentially enables the EA to discover fitter solutions. To enable $z$-axis variability, the genome is extended to include 5 additional parameters in the range [-42,42], each controlling 1/6th of the $z$-axis. After drawing the top layer as before, each new parameter transforms the genome for the next successive $z$-layer by uniformly adding the allele value (capped at the usual bounds), after which it is then drawn in the usual way. Figure~\ref{fig:zreal-res} shows the maximum tip speed achieved by the fittest treatments in each generation. The fittest treatments produced by the GA and surrogate model each generation are shown in Figures~\ref{fig:zga-individuals} and \ref{fig:zmodel-individuals}, respectively. Again, both the GA-only and model-assisted approach design increasingly efficient prototypes. Analysing the final 10 treatments, the average tip speed of the surrogate assisted approach ($M=1217$, $SD=78$, $N=10$) is significantly greater than the GA-only approach ($M=1110$, $SD=41$, $N=10$) using a two-sample {\em t}-test assuming unequal variances, $t(14)=2.14$, $p\le.0018$. Furthermore, the fittest treatment designed by the surrogate assisted approach ($1308rpm$) was greater than the GA-only approach ($1245rpm$) after 100 evaluations. The addition of an extra degree of freedom on the $z$-axis resulted in improved performance for both GA and model-assisted approaches ({\em cf.} Figure~\ref{fig:real-res}).
\begin{figure}[t]
\centering 
\setlength{\fboxsep}{0pt}%
\fbox{\psfig{file=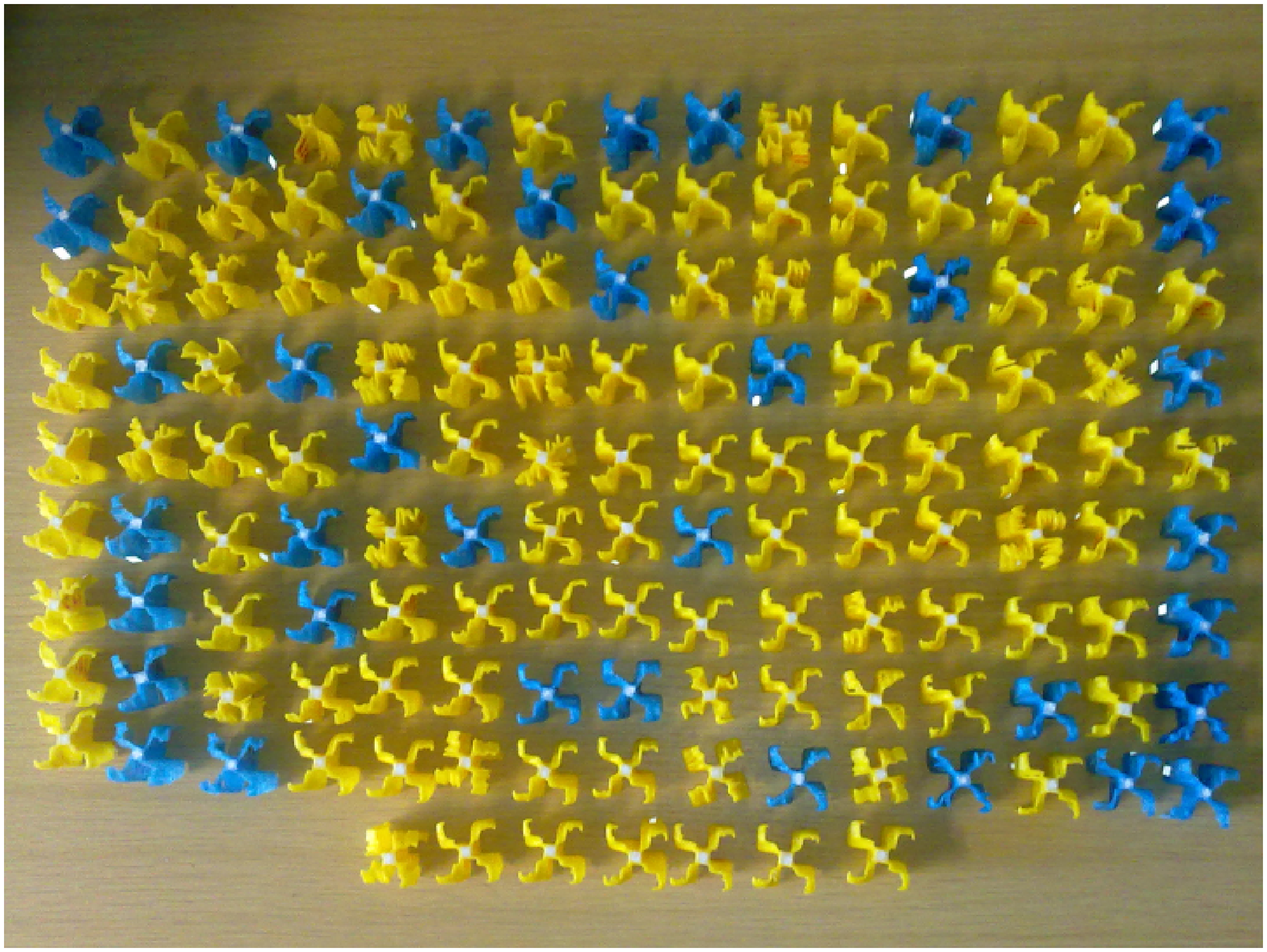,width=5.5in}}
\caption{The fabricated treatments with $z$-variability.}
\label{fig:zall}
\end{figure}
\section{Conclusions}
This paper has shown that EAs are capable of identifying novel and increasingly efficient VAWT designs wherein a sample of prototypes are fabricated by a three-dimensional printer and examined for utility in the real-world (the fabricated designs from one experiment can be seen in Figure~\ref{fig:zall}). The use of a neural network surrogate model was found to reduce the number of fabrications required by an EA to attain higher aerodynamic efficiency (tip speed) of VAWT prototypes, resulting in an important cost reduction. This approach completely avoids the use of three-dimensional computer simulations, with their associated processing costs and modelling assumptions. In this case, three-dimensional CFD analysis was avoided, but the approach is equally applicable to other real-world optimisation problems, for example, those requiring computational structural dynamics simulations. We anticipate that in the future such approaches will yield unusual yet highly efficient designs that exploit characteristics of the environment that are extremely difficult to capture in a simulation.

A vertical-axis wind turbine manufacturer is supporting the development of future work, which will use the power generated by the VAWT prototypes as the fitness computation under various wind tunnel conditions; the exploration of more advanced assisted learning systems to reduce the number of evaluations required; examination of the affect of seeding the population with a given design; investigation of alternative representations that provide more flexible designs including variable number of blades, for example, superquadrics (e.g., \cite{PreenBull:2012}); and the production of full-scale designs. 

If the recent speed and material advances in rapid-prototyping continues, along with the current advancement of evolutionary design, it will soon be feasible to perform a wide-array of automated complex engineering optimisation {\em in situ}, whether on the micro-scale (e.g., drug design), or the macro-scale (e.g., wind turbine design). That is, instead of using mass manufactured designs, EAs will be used to identify bespoke solutions that are manufactured to compensate and exploit the specific characteristics of the environment in which they are deployed. 
\end{document}